\documentclass[10pt]{article} 
\usepackage[preprint]{rlc}

\usepackage{amssymb}            
\usepackage{amsmath}
\usepackage{mathtools}          
\usepackage{mathrsfs}           
\usepackage{graphicx}           
\usepackage{subcaption}         
\usepackage[space]{grffile}     
\usepackage{url}                
\usepackage{algorithm}

\let\classAND\AND
\let\AND\relax
\usepackage{algorithmic}

\let\AND\classAND
\AtBeginEnvironment{algorithmic}{\let\AND\algoAND}

\title{ROER: Regularized Optimal Experience Replay}

\author{Changling Li  \\
    lichan@student.ethz.ch \\
    ETH Zurich
    \And
    Zhang-Wei Hong \\
    zwhong@mit.edu\\
    Massachusetts Institute of Technology 
    \And
    Pulkit Agrawal \\
    pulkitag@mit.edu\\
    Massachusetts Institute of Technology 
    \And
    Divyansh Garg \\
    divgarg@stanford.edu\\
    Stanford University
    \And
    Joni Pajarinen \\
    joni.pajarinen@aalto.fi\\
    Aalto University}

\begin{document}

\maketitle

\begin{abstract}

Experience replay serves as a key component in the success of online reinforcement learning (RL). Prioritized experience replay (PER) reweights experiences by the temporal difference (TD) error empirically enhancing the performance. However, few works have explored the motivation of using TD error. In this work, we provide an alternative perspective on TD-error-based reweighting. We show the connections between the experience prioritization and occupancy optimization. By using a regularized RL objective with $f-$divergence regularizer and employing its dual form, we show that an optimal solution to the objective is obtained by shifting the distribution of off-policy data in the replay buffer towards the on-policy optimal distribution using TD-error-based occupancy ratios. Our derivation results in a new pipeline of TD error prioritization. We specifically explore the KL divergence as the regularizer and obtain a new form of prioritization scheme, the regularized optimal experience replay (ROER). We evaluate the proposed prioritization scheme with the Soft Actor-Critic (SAC) algorithm in continuous control MuJoCo and DM Control benchmark tasks where our proposed scheme outperforms baselines in 6 out of 11 tasks while the results of the rest match with or do not deviate far from the baselines. Further, using pretraining, ROER achieves noticeable improvement on difficult Antmaze environment where baselines fail, showing applicability to offline-to-online fine-tuning. Code is available at \url{https://github.com/XavierChanglingLi/Regularized-Optimal-Experience-Replay}.

\end{abstract}

\section{Introduction}
\label{sec:introduction}


Deep reinforcement learning (RL) have shown wide applications in various domains~\citep{mnih2015human, levine2016end, koert2019learning, li2022energy, hong2024curiosity}. One key factor for its success is the integrated structure of experience replay~\citep{zhang2017deeper}. Experience replay~\citep{lin1992self} allows RL algorithms to use collected experience to compute updates for the current policy. It significantly increases the data efficiency and allows RL to be applied to fields where online data collection is expensive. On the other hand, sampling from experience replay buffer breaks the temporal correlations among experiences and stabilizes the gradient update~\citep{mnih2013playing}. However, past work shows that not all samples are equally informative in updating policy~\citep{katharopoulos2018not}. To enhance the performance, techniques of weighted experience replay~\citep{schaul2015prioritized, kumar2020discor, liu2021regret, sinha2022experience} are proposed to perform importance sampling and shape the distribution of the data in the replay buffer.




Among the proposed reweighting frameworks, prioritized experience replay (PER) is most commonly utilized for its simplicity and empirically good performance~\citep{hessel2018rainbow}. PER attempts to accelerate learning by assigning experiences with the temporal-difference (TD) error to enable higher sampling frequency for transitions with high error. However, PER inherits several shortcomings. First, experience replay reuses experiences from the past iterations to update the current policy. The resulted distribution shift between the data distribution of the replay buffer and the distribution of the current policy can cause incorrect TD error estimations which is detrimental to the performance of PER. On the other hand, it has been empirically shown that staying on policy~\citep{schulman2015trust} or maintaining an on-policy sample distribution can be beneficial to the performance~\citep{sutton2018reinforcement, fu2019diagnosing, novati2019remember}. Second, even though the motivation of using TD error is intuitive, limited works have explored the theoretical foundation~\citep{fujimoto2020equivalence, lahire2021large}.


In this work, we revisit the prioritization scheme and attempt to tackle the aforementioned problems. We provide a new perspective on the TD error prioritization by making connection to the occupancy optimization. We leverage the dual function of the regularized RL objective with $f$-divergence regularizer between off-policy and on-policy data distributions~\citep{Nachum2019AlgaeDICEPG} and show that an optimal solution (occupancy ratio) to the objective is obtained by shifting the off-policy distribution towards the on-policy optimal distribution which results in a TD error prioritization. The form of TD error prioritization is closely associated with the regularized objective which implies that using simple TD error alone may not work best for every RL objectives. On the other hand, introducing regularizer into the objective penalizes TD-error estimation when the distribution of the data from the replay buffer differs too much from the distribution induced by the current policy and thus, gives a smaller priority to mitigate the bias induced by the distribution shift. Together, our derivation provides an alternative perspective on PER and results in a new pipeline of TD-error-based prioritization scheme whose form depends on the choice of the regularizer. Similar to PER, the new framework can be easily integrated with existing RL algorithms by using an additional value network with the regularized objective. 


We specifically focus on KL-divergence as a regularizer and derive its corresponding objective. From this objective, we obtain a new form of prioritized experience replay, the regularized optimal experience replay (ROER). We combine our proposed ROER with Soft Actor-Critic (SAC)~\citep{haarnoja2018soft} algorithm and evaluate on continuous control MuJoCo and DM control benchmark tasks. ROER outperforms baselines in 6 out of 11 tasks while the rest match with or do not deviate far from the best performance. Especially, ROER shows performance improvements on environments where PER and LaBER~\citep{lahire2021large} fails. Further evaluation on the value estimations shows that the performance improvement of ROER attributes to the more accurate value estimation by mitigating the underestimation bias of SAC with double critics~\citep{li2021balancing, zhou2022revisiting} and thus ROER can obtain or converge to the optimal solutions much faster than baselines. Further, we consider the setting of online with pretraining and ROER achieves noticeable improvement on difficult Antmaze environment whereas the baselines fail, showing the applicability to offline-to-online fine-tuning.   

\section{Preliminaries}
\textbf{Online RL.} Online reinforcement learning concerns optimizing an agent's policy in a Markov decision process (MDP) \citep{puterman2014markov}. The MDP is defined by a tuple $\mathcal{M} = (\mathcal{S}, \mathcal{A}, P, r, \rho_0, \gamma)$ where $\mathcal{S}$ and $\mathcal{A}$ represent the state space and action space respectively, $P(s'|s, a)$ denotes the dynamic model, $r(s, a)$ the reward function, $\rho_0$ the initial state distribution, and $\gamma \in (0, 1)$ the discount factor. The agent's behavior is described by its policy $\pi: \mathcal{S} \rightarrow \Delta_{\mathcal{A}}$. The performance of a given policy can be measured by the state-action value function $Q^{\pi}(s, a) = \mathbb{E}[\sum^{\infty}_{t=0}\gamma^t r(s_t, a_t)|s_0=s, a_0=a, s_{t+1} \sim P(\cdot|s_t, a_t), a_t\sim \pi(\cdot|s_t)]$. The corresponding value function is $V^{\pi}(s):= \mathbb{E}[Q^{\pi}(s,a)|a\sim \pi(\cdot|s)]$. The goal is to learn a policy that maximizes the $\gamma$-discounted expected cumulative return \citep{sutton2018reinforcement}: 
\begin{align}
\max_\pi \mathcal{J}_P(\pi) := (1-\gamma)\mathbb{E}_{s_0 \sim \rho_0, a_0 \sim \pi(\cdot|s_0)}[Q^\pi(s_0, a_0)]    
\end{align}
For a fixed policy $\pi$, we can rewrite the expected return in terms of its state-action distribution \citep{wang2007dual,puterman2014markov} as 
\begin{align}
    \max_\pi \mathcal{J}_D(\pi):=\mathbb{E}_{(s,a)\sim d^\pi}[r(s, a)]
\end{align}
where  $d^\pi(s,a) = (1-\gamma) \sum^{\infty}_{t-0} \gamma^t \text{Pr}[s_t=s, a_t=a|s_0 \sim \rho_0, a_t\sim \pi(\cdot|s_t), s_{t+1}\sim P(\cdot|s_t, a_t)]$.

In actor critic methods, one alternates between updating policy $\pi$ (the actor) and Q-approximator $Q_\theta$ (the critic)\citep{konda1999actor}. The policy updates according to the policy gradient theorem \citep{sutton1999policy} as 
\begin{equation}
    \frac{\partial}{\partial \pi}\mathcal{J}_P(\pi) = \mathbb{E}_{(s, a)\sim d^{\pi}}[Q^{\pi}(s,a)\nabla \text{log}\pi(a|s)]
\end{equation}
The critic is learned via TD learning based on Bellman equation~\citep{bellman1966dynamic} $\mathcal{B^\pi}Q^{\pi} := r(s, a) + \gamma \mathbb{E}_{s',a'}Q^{\pi}(s', a')$ where $\mathcal{B^\pi}$ denotes the expected Bellman operator. Given some experience replay buffer $\mathcal{D}$ collected in the same MDP but by potentially different policies, the Q-approximator is learned via a variation of the following form 
\begin{equation}
    \text{min}_{Q_\theta}\mathcal{J}(Q_\theta) := \frac{1}{2}\mathbb{E}_{(s,a)\sim \mathcal{D}} [(\mathcal{B^\pi}Q_{\theta}-Q_{\theta})(s,a)^2].
\end{equation}
In practice, we generally cannot access the true target value $B^\pi Q_{\theta}$ and thus, we use an estimation $\hat{B}^\pi Q_{\theta}$ to fit $Q_{\theta}(s,a)$.

\textbf{Prioritization in Experience Replay.}
Prioritization in experience replay applies weighted sampling to the experiences by assigning weights to individual state-action which is equivalent to the weighted objective. We define the weight for a experience with state $s$ and action $a$ as $w(s,a)$ which is positive. Then, under the sampling distribution $d \in P(\mathcal{S} \times \mathcal{A}) $, we have the weighted learning objective:
\begin{equation}\label{eq: weighted Q}
    \text{min}_{Q_\theta}\mathcal{J}(Q_\theta) := \frac{1}{2}\mathbb{E}_{d} [w(s,a)(\mathcal{B^\pi}Q_{\theta}-Q_{\theta})(s,a)^2].
\end{equation}
In practice, $w(s,a)$ can be in various forms such as likelihood~\citep{sinha2022experience}. TD error is the most commonly considered and it forms prioritized experience replay (PER) which samples transitions proportional to their TD errors~\citep{schaul2015prioritized}. We denote TD error as $\delta$ and at time step $t$, it is defined as 
\begin{equation}
    \delta_t = r_{t+1} + \gamma V(s_{t+1}) - V(s_t).
\end{equation}
Even though PER shows heuristically good results, the motivation of using TD error is under explored. In addition, using the TD error estimated by the Q-function induced by the current policy can be sub-optimal as the estimation of TD error can be inaccurate especially on the states that are infrequently visited under the current policy.  


\section{Experience Prioritization as Occupancy Optimization}
\label{methods}

The goal of using prioritization is to accelerate learning and obtain an optimal policy which induces an optimal on-policy distribution. We reverse this process and motivate our formulation by the problem of obtaining an optimal policy by finding the optimal on-policy distribution $d^*$, given access to an off-policy distribution $d^\mathcal{D}$.
Here, $d^*$ is  unknown and  we assume to only have samples from $d^\mathcal{D}$ which is the distribution of the experience replay buffer. For an MDP with a reward function $r$, there exists a unique $d^*$. We consider the following regularized objective with an $f$-divergence regularizer to include $d^\mathcal{D}$~\citep{nachum2019dualdice}
\begin{equation}
\label{eq:obj_reg}
\max _{d^\mathcal{D}} \mathcal{J}_{D,f}(d^*, d^\mathcal{D}):=\mathbb{E}_{(s, a) \sim d^*}[r(s, a)]-\beta D_f\left(d^* \| d^{\mathcal{D}}\right)
\end{equation}

where $\beta > 0$ and $D_f$ denotes the $f$-divergence induced by a convex function $f$:
\begin{align}
    D_f(d^*||d^\mathcal{D}) = \mathbb{E}_{(s,a)\sim d^\mathcal{D}}[f(w_{*/\mathcal{D}}(s,a))]
\end{align}
where $w_{*/\mathcal{D}}:= \frac{d^*(s,a)}{d^\mathcal{D}(s,a)}$. The regularizer encourages conservative estimation and serves as a penalty when the off-policy distribution deviates too much from the on-policy distribution. Note that the strength of regularization can be controlled by $\beta$. The above objective $\mathcal{J}$ is maximized for $d^\mathcal{D} = d^*$, where it becomes the unconstrained RL problem. 


The regularized objective (Equation~\ref{eq:obj_reg}) can be transformed to the following dual problem~\citep{Nachum2019AlgaeDICEPG, nachum2020reinforcement}. Let $x: S \times A \rightarrow \mathbb{R}$. We have the dual function of $\mathcal{J}$:
\begin{align}
 \tilde{\mathcal{J}}_{D, f}(d^*, d^\mathcal{D}) = \min_x \mathbb{E}_{(s, a) \sim d^*}[r(s, a)]+\beta \mathbb{E}_{(s, a) \sim d^D}[f_*(x(s, a))] - \beta \mathbb{E}_{(s, a) \sim d^*}[x(s, a)]]
\end{align}

where $f_*$ is the convex conjugate of $f$. Note that the optimal $x^*(s,a)$ w.r.t the dual objective satisfies $f'_*(x^*) = d^*/d^\mathcal{D}$. We apply the change of variable. Let $Q(s,a)- \gamma V^*(s')= -\beta x(s,a) + r(s,a)$ where $Q(s,a)$ is a fixed point of a variant of Bellman equation~\citep{nachum2019dualdice} and $\gamma V^*(s') + r(s,a) = \mathcal{B^*}Q(s,a)$. We obtain the new objective which is independent of $d^*$:
\begin{equation}
    \tilde{\mathcal{J}}_{D, f}(d^*, d^\mathcal{D}) = \min_Q \ \beta \cdot \mathbb{E}_{(s, a) \sim d^\mathcal{D}}\left[f_*\left((\mathcal{B}^* Q(s, a) - Q(s, a))/\beta\right)\right] + (1-\gamma)  \mathbb{E}_{s_0 \sim \mu_0, a_0 \sim \pi^*(s_0)}\left[Q\left(s_0, a_0\right)\right]    
\end{equation}
Using $\delta_Q := \mathcal{B}^* Q(s, a)-Q(s, a)$ to denote the TD error, we obtain a solution $Q^*$ to the objective satisfying:
\begin{equation}
    f'_*(\delta_{Q^*}/\beta) = d^*/d^\mathcal{D}, 
\end{equation}
which gives the TD-error based occupancy ratio between the optimal distribution and the current distribution. We point out two key observations:
\begin{itemize}
    \item Using the property of convex conjugate: $f'_*(f'(x)) = x$ and $f''(x) \geq 0$, we can rewrite $
    \mathcal{B}^* Q^*(s, a)-Q^*(s, a) = \beta f'(d^*/d^\mathcal{D})$. By absorbing the term $\beta \mathbb{E}_{(s, a) \sim d^*}[x(s, a)]]$ of the dual objective into the reward, we have that $Q^*$ is the optimal Q-function for the augmented reward $\tilde{r} = r - \beta f'(d^*/d^\mathcal{D})$.
    \item When $d^* = d^\mathcal{D}$, as $f'(1) = 0$,  $Q^*$ is the optimal Q-function to the reward $r$ and solves the unregularized RL problem of maximizing $r$.
\end{itemize}
Thus, in theory, the above problem has a unique saddle point solution where $d^\mathcal{D} = d^*$ and $Q^*$ is the optimal Q-function, which can be found by shaping $d^\mathcal{D}$ towards $d^*$ using the following weighting formulation:
\begin{equation}
    d^* = f'_*(\delta_{Q^*}/\beta) \cdot d^\mathcal{D}.     
\end{equation}
We include the details of derivation in Appendix~\ref{Apdix: proof}. In practice, we have a changing distribution of $d^\mathcal{D}$ for online reinforcement learning due to the collection of the new data and update to the policy. However, we solve the optimization problem in many steps. We show that the distribution can still asymptotically converge to the optimal $d^*$ by empirically showing that our proposed method mitigates the value underestimation bias of Soft-Actor Critic with double q-learning and converges to the true value in section~\ref{sec: value estimation}.

\section{Regularized Optimal Experience Replay}

In this section, we discuss our choice of 
using Kullback-Leibler (KL) divergence as the regularizer and proceed to the practical implementation of the prioritization scheme with the KL-divergence regularizer which forms our proposed method, the regularized optimal experience replay (ROER). Other forms of $f$-divergence can also be suitable candidates and we provide further discussions in Appendix~\ref{Apdix: other divergence}.  

\subsection{KL Divergence as Regularizer}
$f$-divergence consists of numerous forms and past works have explored the application of it in the policy update rules of RL~\citep{belousov2017f, kumar2020conservative}. Particularly, many works focus on KL-divergence as it improves the efficiency and performance of RL algorithms such as Trust Region Policy Optimization~\citep{schulman2015trust} and Maximum a Posteriori Policy Optimization~\citep{abdolmaleki2018maximum}. Theoretical exploration also shows the advantage of KL regularization~\citep{vieillard2020leverage}. Thus, we consider KL-divergence as the regularizer in our formulation for it penalizing the off-policy distribution being too far from the on-policy distribution and the later-on derived objective. 

Recall that the function of KL-divergence has the form $f(x) = x\log(x)$ and its convex conjugate has the form $f_*(y) =  e^{y} - 1$. Let $y=(\mathcal{B}^* Q(s, a) - Q(s, a))/\beta$, we follow the derivation in section~\ref{methods} and obtain the following objective:
\begin{equation}
    \min_Q \ \mathbb{E}_{(s, a) \sim d^\mathcal{D}}\left[e^{\left(\mathcal{B}^* Q(s, a) - Q(s, a)\right)/\beta}\right] - \mathbb{E}_{(s, a, s') \sim d^\mathcal{D}}\left[\mathcal{B}^*Q(s,a) - Q(s,a)\right]-1
\end{equation}
We note that this objective is reminiscent to the loss function of Extreme-Q learning~\citep{garg2023extreme} which leverages Extreme Value Theory to avoid computing Q-values using out of distribution actions and thus, mitigate the estimation error. This allows for obtaining a more accurate TD error for priority calculation. We note that our method differs from extreme q-learning as we only uses this loss to obtain TD error to shape the data distribution towards an optimal on-policy distribution. Using this objective, the occupancy ratio has the form
\begin{equation}
    d^*/d^\mathcal{D} = f_*'(\delta_Q/\beta) = e^{\delta_Q/\beta}
\end{equation}
which gives our proposed regularized optimal experience replay formulation.

\subsection{Practical Implementation}
\begin{algorithm}[htp]
\caption{Actor Critic with Regularized Optimal Experience Replay}
\begin{algorithmic}[1] 
\label{alg:coer}
\STATE Initialize $Q_\theta$, $\pi_\psi$, value network $V_\phi$, training start step $\tau$
\STATE Let $\mathcal{D}$ be the empty replay buffer or filled with offline data with $d(s,a)=1$
    \FOR{step $t$ in ${1, ..., N}$}
        \STATE Update (s, a, r, s') to $\mathcal{D}$ with $d(s,a)=1$
        \IF{t $\geq$ $\tau$}
            \STATE Update $d(s,a)$ with $d'$ from Eq. \ref{eq: update scheme}\
            \STATE Train $Q_\theta$ with $J(Q_\theta)$ from Eq. \ref{eq: weighted Q} using $d(s,a)$ as $w(s,a)$
            \STATE Train $V_\phi$ with $\mathcal{L}(V)$ from Eq. \ref{eq: value loss}
            \STATE Update $\pi_\psi$
        \ENDIF 
    \ENDFOR
\STATE \textbf{return} $Q^*, \pi^{*}$
\end{algorithmic}
\end{algorithm}
Note that the form of occupancy ratio is derived from a regularized objective which can be different from the objective of the applied algorithm. For smooth integration to the existing algorithms, we propose to incorporate a separate value network using the regularized objective for TD error estimation and priority calculation. The above KL divergence gives the value network the following objective of the ExtremeV loss~\citep{garg2023extreme} 
\begin{align}
    \mathcal{L}(\mathcal{V}) = \mathbb{E}_{(s, a) \sim d^\mathcal{D}}\left[e^{\left(\mathcal{B}^* Q(s, a) - V(s)\right)/\beta}\right] - \mathbb{E}_{(s, a, s') \sim d^\mathcal{D}}\left[ Q(s,a) - V(s)\right]-1.
    \label{eq: value loss}
\end{align}

We then use the TD error obtained from the value network to calculate the priority. Since the distribution of $d^\mathcal{D}$ is changing, we consider a stable convergence and solve the optimization problem in many steps. We introduce a convergence parameter $\lambda$ and formulate the following priority update function 
\begin{equation}
    d' = [\lambda e^{\delta_{Q^*}/\beta}  + (1-\lambda)] \cdot d^\mathcal{D}  \text{ with } \lambda \in (0, 1].
    \label{eq: update scheme}
\end{equation}
where $d^{\mathcal{D}}$ is the current priority of the samples and $d'$ is the updated priority (used as $w$). In an online setting, we start by assigning each sample in replay buffer $\mathcal{D}$ with priority $d=1$ and use the above update function to update the priority after each Q-iteration step. The loss temperature $\beta$ here controls the scale of TD-error and thus, the scale of the priority. We note that exponential function is sensitive to outliers. Thus, we use mean normalization and clip the exponential of TD error and the priority to control the range and avoid outliers. The general procedure of our approach is summarized in Algorithm~\ref{alg:coer} and more implementation details are listed in Appendix~\ref{Apdix: experiments}. The proposed priority update function slowly improves the current distribution $d'$ towards the optimal policy distribution $d^*$, and ultimately maximizes the objective $\mathcal{J}$. 





\section{Experimental Evaluation}

We combine our proposed prioritization scheme ROER with Soft-Actor Critic~\citep{haarnoja2018soft} algorithm for evaluation. We compare our method with two state-of-art prioritization schemes namely uniform experience replay (UER) and the initial TD error prioritized experience replay (PER)~\citep{schaul2015prioritized}, and one additional baseline namely large batch experience replay (LaBER)~\citep{lahire2021large} across a wide set of MuJoCo continuous control tasks interfaced through OpenAI Gym~\citep{brockman2016openai} and DM Control tasks~\citep{tunyasuvunakool2020dm_control} in an online setting. Additionally, we consider a suite of more difficult environment Antmaze with pretraining using the data from D4RL~\citep{fu2020d4rl} to show that ROER can achieve good performance in settings where both UER and PER fail. To allow for reproducibility, we use the original set of tasks without modification to the environment or rewards. For a fair comparison between baselines and our approach, our implementations are all based on JAXRL~\citep{jaxrl}. 

Compared to the initial PER, even though our proposed method ROER has four more hyperparameters namely the architecture of value network, loss temperature ($\beta$), Gumbel loss clip (Grad Clip), and maximum exponential of TD-error clip (Max Exp Clip), we note that $\beta$ and Grad Clip are not new and they come from the objective of Extreme Q-Learning. Grad Clip is shown to affect the results lightly and the value network can use the default parameters as the critic network. A set of values works well for multiple environments. We provide more details of implementation, ablations and hyperparameters in Appendix \ref{Apdix: experiments}.




\subsection{Online}

\begin{table}[h]
\begin{center}
\begin{tabular}{l|llll}
\hline
Env & SAC & SAC+PER & SAC+LaBER & SAC+ROER (ours) \\ \hline
Ant-v2 & $1153.1 \pm 335.5$ & $1654.1 \pm 342.9$ & $1006.0 \pm 546.0$ & $\mathbf{2275.5} \pm 598.6$ \\
HalfCheetah-v2 & $9017.4 \pm 172.5$  & $9240.4 \pm 276.5$ & $7962.8 \pm 304.5$ & $\mathbf{10695.5} \pm 183.4$ \\
Hopper-v2 & $2813.0 \pm 481.2$ & $2937.7 \pm 334.3$ & $2330.8 \pm 514.3$ &  $\mathbf{3010.2} \pm 299.0$ \\
Humanoid-v2 & $5026.8 \pm 154.1$ & $4993.4 \pm 198.0$ & $5000.9 \pm 319.5$ & $\mathbf{5257.0} \pm153.2$ \\
Walker2d-v2 & $\mathbf{4344.3} \pm 177.7$ & $4003.9 \pm 318.7$  & $4033.1 \pm 375.7$ & $4328.5 \pm 311.4$ \\ \hline
Fish-swim & $247.7 \pm 59.6$ & $234.6 \pm 63.6 $ & $178.3 \pm 49.9$ & $\mathbf{301.9} \pm 54.9$ \\
Hopper-hop & $134.4 \pm 34.2$ & $\mathbf{147.2} \pm 31.3$ & $146.7 \pm 29.8$ & $125.7 \pm 35.2$ \\
Hopper-stand & $521.1 \pm 120.1$ & $384.7 \pm 94.9$ & $475.5 \pm 111.0$ & $\mathbf{798.5} \pm 89.2$ \\
Humanoid-run & $130.3 \pm 21.7$ & $116.3 \pm 18.7$ & $\mathbf{144.8} \pm 18.1$ & $137.3 \pm 12.3$ \\ 
Humanoid-stand & $733.4 \pm 53.9$ & $765.0 \pm 38.8$ & $\mathbf{827.8} \pm 40.9$ & $691.6 \pm 57.8$ \\
Quadruped-run & $761.2 \pm 89.4$ & $606.2 \pm 114.7$ & $\mathbf{796.3} \pm 82.6$ & $772.1 \pm 77.7$ \\\hline
\end{tabular}
\end{center}
\caption{Average evaluation performance attained over the last 10 evaluations over 1 million time steps for MuJoCo and DM Control tasks. Average performance and $95\%$ confidence interval ($\pm$) are attained over 20 random seeds. Maximum average value for each task is highlighted as bold}
\label{Tab:online_main_results}
\end{table}

In the online setting, the empirical results demonstrate that our proposed ROER outperforms state-of-the-arts on 6 out of 11 continuous control tasks in terms of the average evaluation while do not deviate far from the baselines for the rest 3 environments as shown in Table~\ref{Tab:online_main_results}. We find that ROER with SAC achieves noticeable improvement on HalfCheetah-v2, Ant-v2, Humanoid-v2 from MuJoCo and Fish-Swim, Hopper-stand from DM Control. Especially for Hopper-stand environment, our proposed ROER with SAC nearly doubles the performance of UER or PER with SAC. We attribute the improvements to the more accurate TD error estimation using a separate value network with divergence regularized objective and the associated priority update form. Within the five under-performed tasks, ROER obtains a similar performance as the UER in Walker2d-v2 and outperforms PER and LaBER.

In contrast, PER only shows slight improvement on limited number of continuous control tasks compared to UER including HalfCheetah-v2, Hopper-v2, Hopper-hop, and Humanoid-stand. In many tasks, PER even worsens the performance such as Walker2d, Hopper-stand, Humanoid-run and Quadruped-run whereas our proposed method can maintain a similar or achieve much better performance. We consider the reasons for PER failing to be the biased priority induced by the inaccurate TD error estimation and the less stable priority update scheme. We note that LaBER achieves better results in Humanoid-run, Humanoid-stand and Quadruped-run but in the cost of much longer training time due to the larger batch required by the algorithm. The hyperparameter selection for LaBER can be found in Appendix \ref{Apdix: experiments}.

We notice that all three prioritization schemes under perform in Ant-v2. Our proposed ROER with SAC, even though achieves higher average performance in Ant-v2, has a very large confidence interval. This requires additional tuning to hyperparameters of SAC and training steps. We keep our current results for a fair comparison across tasks. Evaluation curves and additional discussion of results can be found in Appendix~\ref{Apdix: additional results}.

\subsection{Value Estimation Analysis}



\label{sec: value estimation}
\begin{figure}[h] 
\begin{center}
\begin{subfigure}{0.325\textwidth}
\includegraphics[width=\linewidth]{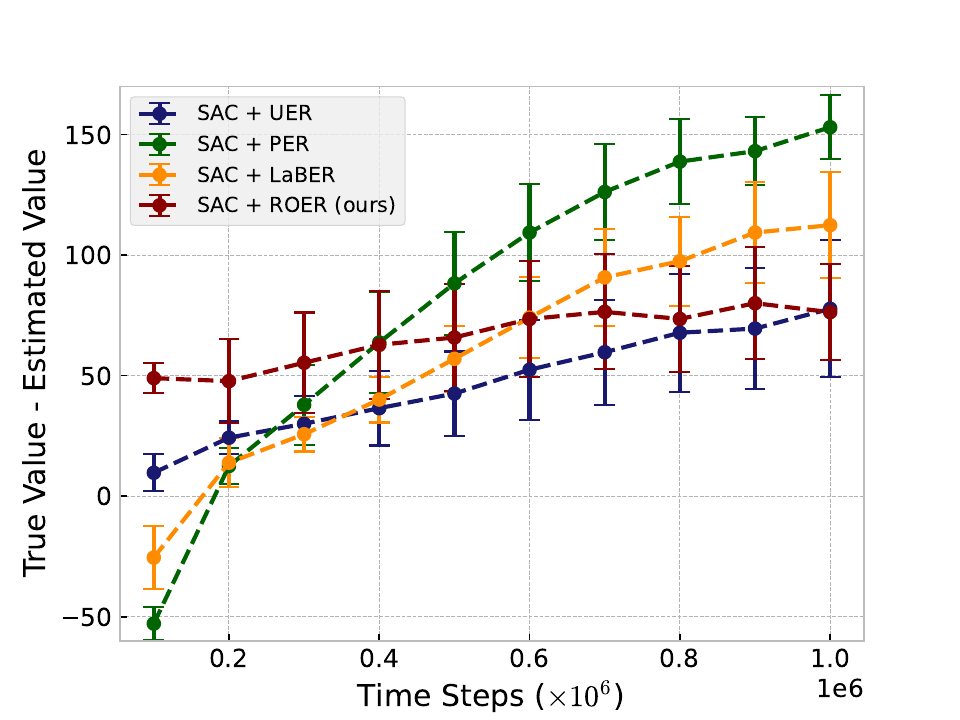}
\caption{Ant-v2}\label{fig:ant_value_est}
\end{subfigure}
\begin{subfigure}{0.325\textwidth}
\includegraphics[width=\linewidth]{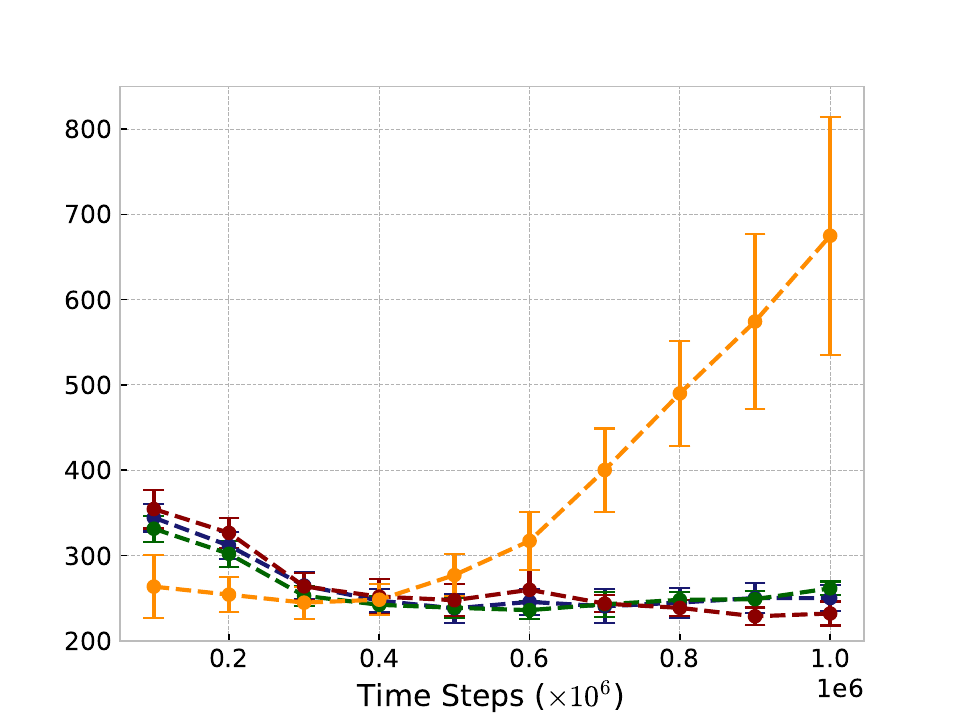}
\caption{HalfCheetah-v2}\label{fig: halfcheetah_value_est}
\end{subfigure}
\begin{subfigure}{0.325\textwidth}
\includegraphics[width=\linewidth]{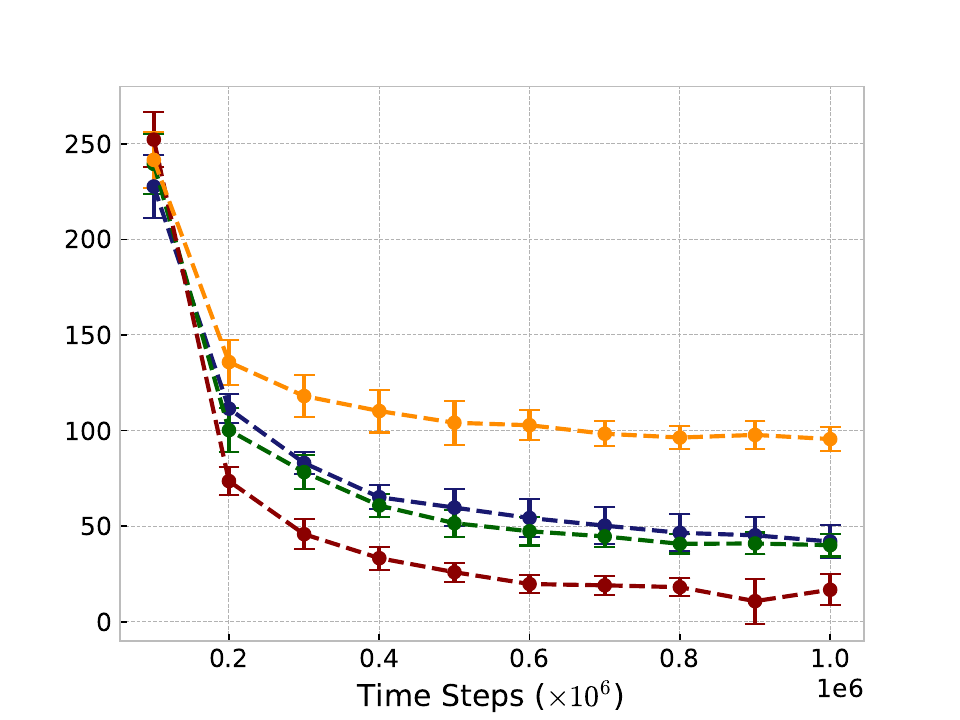}
\caption{Hopper-v2}\label{fig:hoppper_value_est}
\end{subfigure}

\begin{subfigure}{0.325\textwidth}
\includegraphics[width=\linewidth]{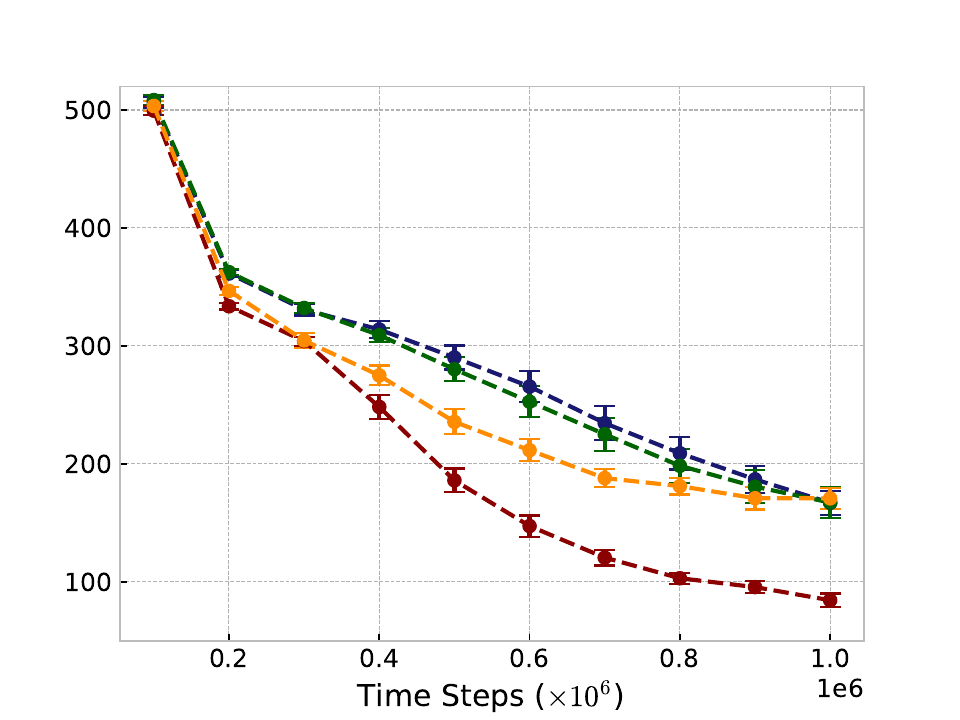}
\caption{Humanoid-v2}\label{fig:humanoid_value_est}
\end{subfigure}
\begin{subfigure}{0.325\textwidth}
\includegraphics[width=\linewidth]{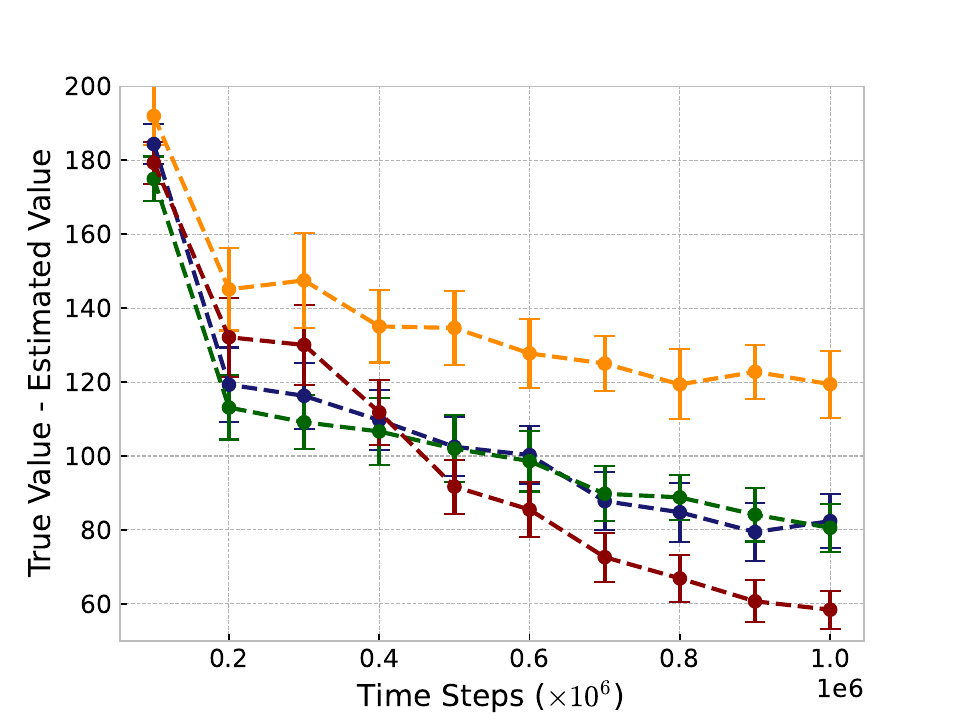}
\caption{Walker2d-v2}\label{fig:walker2d_value_est}
\end{subfigure}
\caption{Measuring underestimation bias in the value estimates of SAC, SAC with PER and SAC with ROER of continuous control tasks in MuJoCo by the difference between the true values and the value estimates. True value is obtained by Monte Carlo returns. Value estimates and true values are averaged over 20 random seeds and the error bar represents 95\% confidence interval.} \label{fig:value_estimation}
\end{center}
\end{figure}

The better performance can be empirically confirmed by the faster convergence to the true value. Besides the derivation that shows our proposed prioritization scheme results in the optimal solution, ROER also demonstrates empirically more accurate value estimation and faster convergence than the baselines. We compare the value estimation with the true value for each algorithm trained online over tasks in MuJoCo as shown in Fig.~\ref{fig:value_estimation}. SAC with double critics tends to underestimate the value~\citep{li2021balancing, zhou2022revisiting}. Compared to the baselines, ROER reduces the underestimation biases and converge to the true values much faster especially in Hopper-v2, Humanoid-v2 and Walker2d-v2. We note that ROER reaches the true value in reward saturated cases such as Hopper-v2 while the baselines still show the underestimation bias. This result serves as an empirical evidence that our proposed prioritization scheme reshapes the replay buffer towards the optimal on-policy distribution and results in the optimal $Q^*$ which is the solution to the objective.

\subsection{Online with Pretraining}
\begin{figure}[ht] 
\begin{subfigure}{0.24\textwidth}
\includegraphics[width=\linewidth]{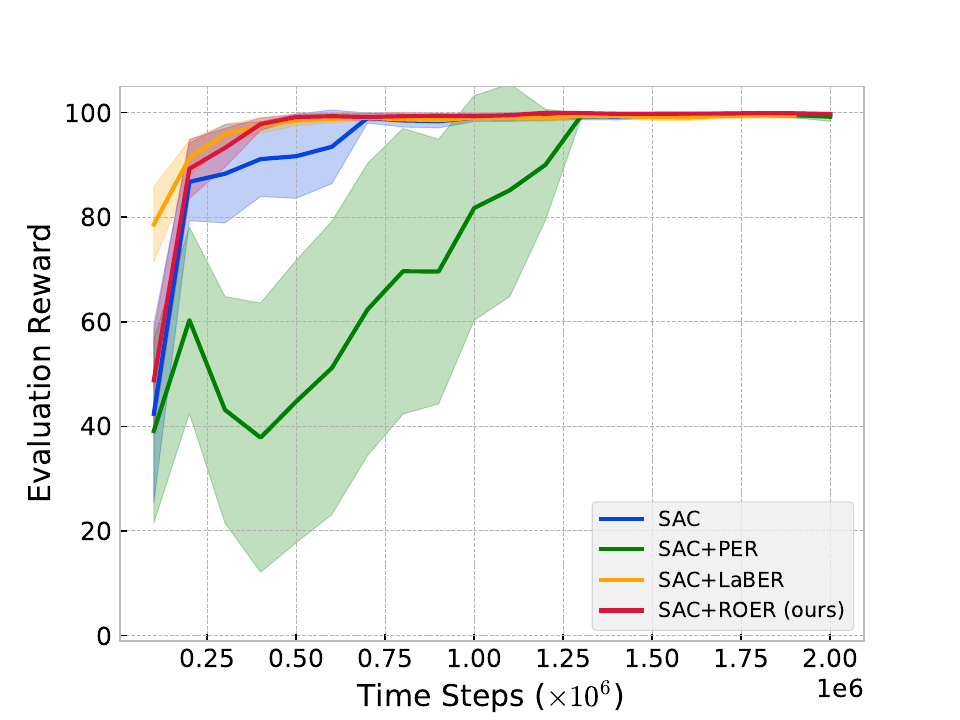}
\caption{Umaze-V2} \label{fig:antmaze-umaze}
\end{subfigure}
\begin{subfigure}{0.24\textwidth}
\includegraphics[width=\linewidth]{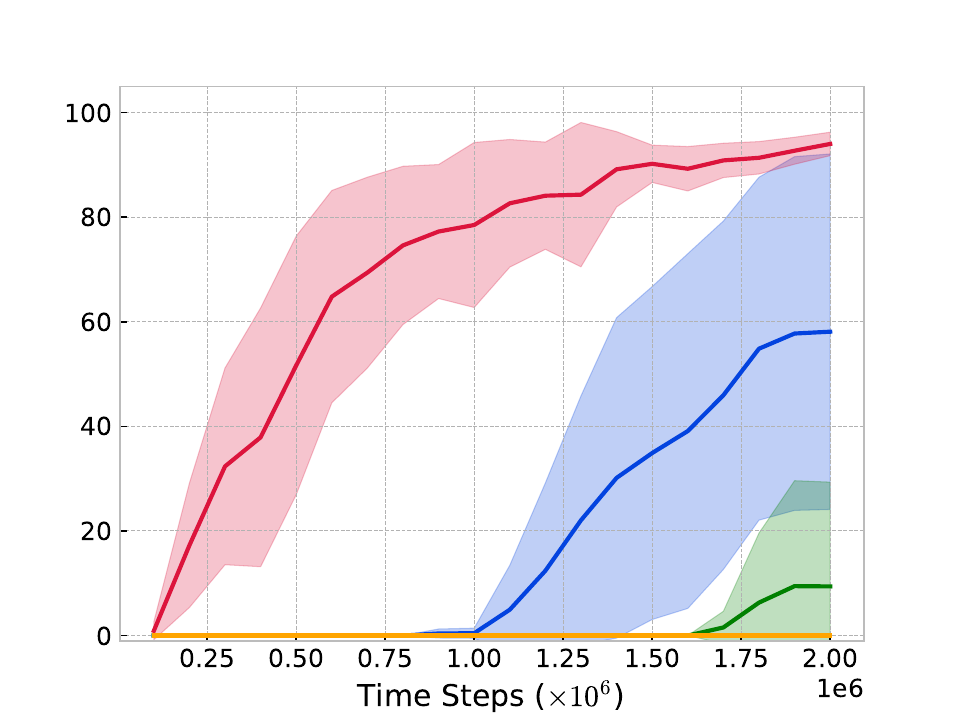}
\caption{Umaze-Diverse-V2} \label{fig:antmaze-umaze-diverse}
\end{subfigure}
\begin{subfigure}{0.24\textwidth}
\includegraphics[width=\linewidth]{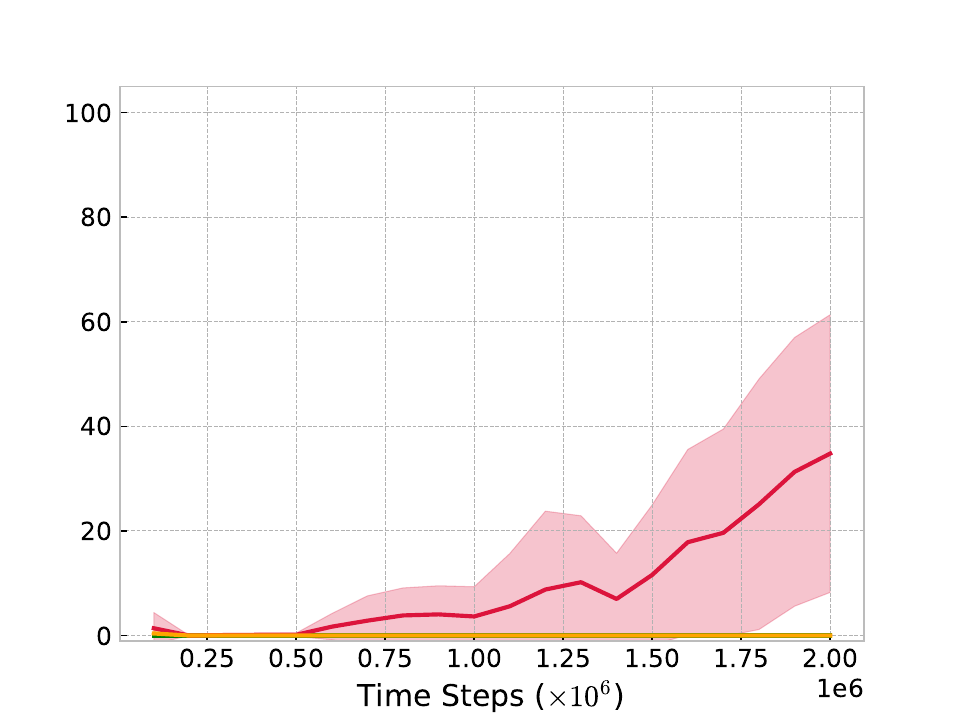}
\caption{Medium-Play-V2} \label{fig:antmaze-med-play}
\end{subfigure}
\begin{subfigure}{0.24\textwidth}
\includegraphics[width=\linewidth]{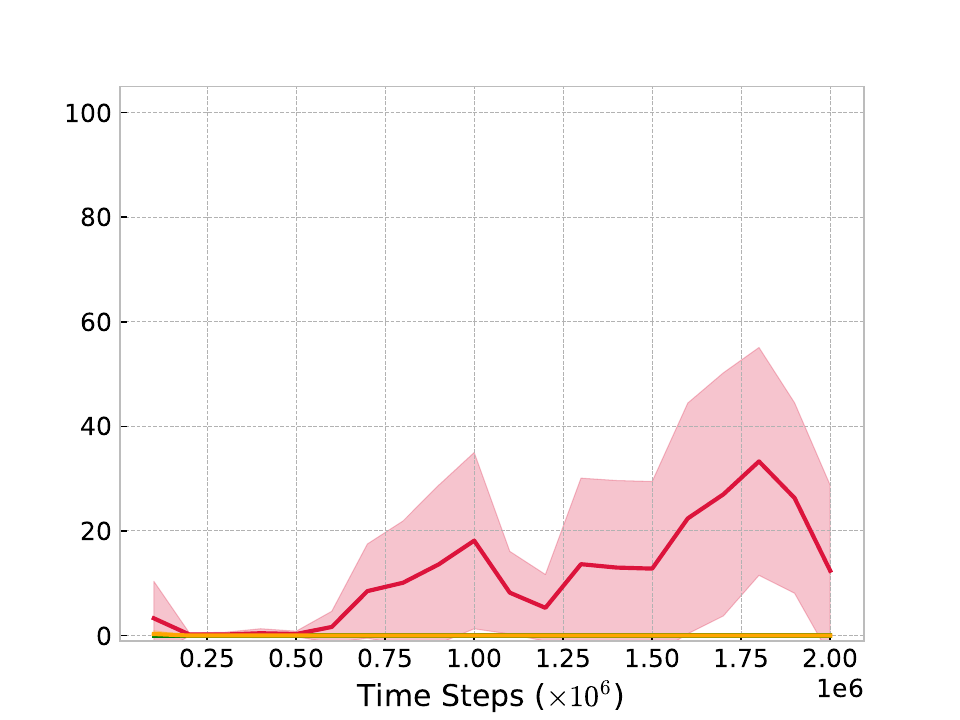}
\caption{Medium-Diverse-V2} \label{fig:antmaze-med-diverse}
\end{subfigure}
\caption{Learning curves for the Antmaze tasks in Gym-Robotics with data from D4RL. Curves are averaged over 10 random seeds, where the shaded area represents the standard error of the average evaluation.} \label{fig:antmaze}
\end{figure}

Our proposed ROER prioritization scheme can benefit from pretraining using offline data and show significant performance improvement over more difficult environment Antmaze-Umaze and Antmaze-Medium as revealed in Table~\ref{tab:antmaze}. We recognize that using the average performance over the last 200 evaluations may not be the most suitable metric here due to the sparsity of rewards and the difficulty of the environments. Thus, we also include the learning curves to illustrate the results as in Fig.~\ref{fig:antmaze}. We found that SAC with ROER can obtain good performance at a very early stage in Antmaze-Umaze environment using both Antmaze-Umaze-v2 dataset and Antmaze-Umaze-diverse-v2 dataset. Especially with Antmaze-Umaze-diverse-v2 dataset, SAC with ROER achieved a significantly better performance compared to the state-of-art prioritization schemes while PER and LaBER are shown to be detrimental to the learning process. We also note that for a more difficult environment Antmaze-Medium, our proposed method can obtain rewards at an early stage and shows improvement over training steps as in Fig.~\ref{fig:antmaze-med-play} and Fig.~\ref{fig:antmaze-med-diverse}. In contrast, SAC with UER, SAC with PER and SAC with LaBER completely fail to obtain any reward signal. This implication is crucial to improve training efficiency and safety by using offline data and correct the distribution to obtain a good performance online. It shows the potential applicability of our method in offline-to-online finetuning. 

\begin{table}[h]
\begin{center}
\begin{tabular}{l|llll}
\hline
Env & SAC & SAC+PER & SAC+LaBER & SAC+ROER (ours)\\ \hline
antmaze-umaze-v2 & $99.6 \pm 0.4$ & $99.6 \pm 0.5$ & $99.45 \pm 0.5$ &$\mathbf{99.9} \pm 0.2$ \\
antmaze-umaze-diverse-v2 & $57.8 \pm 22.7$ & $9.5 \pm 13.5$ &  $0.0$ &  $\mathbf{92.7} \pm 1.9$ \\
antmaze-medium-play-v2 & $0.0$ & $0.0$ & $0.0$ & $\mathbf{31.3} \pm 17.3$ \\
antmaze-medium-diverse-v2 & $0.0$ & $0.0$ & $0.0$ & $\mathbf{26.3} \pm 14.0$ \\\hline
\end{tabular}
\end{center}
\caption{Average evaluation performance attained over the last 200 evaluations over $2e6$ time steps after Antmaze environments. Average performance and $95\%$ confidence interval $(\pm)$ are attained over 10 random seeds.}
\label{tab:antmaze}
\end{table}


\section{Related Work}
Our approach builds upon regularized RL objective and weighted experience replay. 

\textbf{Regularized RL.} 
Regularization is commonly utilized in offline reinforcement learning to constrain the behavior policy and action selection~\citep{kumar2020conservative, wu2019behavior, kumar2019stabilizing}. Other works have considered regularized Q-function of the behavior policy~\citep{shi2023offline} and state-action value offset~\citep{kostrikov2021offline}. Adapting such regularizers in online setting can achieve more stable performance~\citep{fujimoto2019off, schulman2015trust}. Maximizing the regularizer as a way to encourage exploration also shows improvement in performance and forms the framework of max entropy RL~\citep{Ziebart2008MaximumEI, haarnoja2017reinforcement, haarnoja2018soft}. Our work builds upon the line of work that utilizes the dual function of the regularized objective~\citep{belousov2017f, nachum2020reinforcement, Nachum2019AlgaeDICEPG} which allows to express the max-return optimization by an expectation over an arbitrary behavior-agnostic and off-policy data distribution. We extend this approach and formulate the prioritization scheme that allows the data distribution in replay buffer gradually converge to the optimal distribution which gives the optimal Q-function. Theoretical analysis of the regularized RL shows that despite its non-convexity, this problem has zero duality gap and can be solved exactly in the dual domain~\citep{geist2019theory, neu2017unified, paternain2019constrained}.

\textbf{Weighted experience replay.}  
Experience replay is crucial to the success of deep RL for improving the data efficiency by using off-policy data~\citep{lin1992self, hessel2018rainbow}. Various frameworks have been proposed to change the sampling strategy to achieve superior performance than uniform sampling. Prioritized experience replay (PER) weights the experiences by their TD errors and shows empirical improvement when applying to deep RL~\citep{schaul2015prioritized, fujimoto2020equivalence}. However, few works have explored the theoretical motivation of using TD-error based reweighting scheme. \citet{lahire2021large} suggest that PER can be considered as an importance sampling scheme using approximated per-sample gradient norms and prioritizing stochastic gradient descent variance reduction. Our work, on the other hand, uses dual function of the regularized RL objective to provide an alternative perspective on TD-error-based prioritization. Other considerations of prioritization scheme include loss value~\citep{hessel2018rainbow}, accuracy of the TD-error estimation~\citep{sinha2022experience}, regret minimization~\citep{liu2021regret}, and leveraging neural network for experience selection~\citep{zha2019experience}. We note that~\citet{kumar2020discor} shares similarity to our work in correcting the replay buffer towards optimal distribution. However, they consider optimizing corrective feedback while our work builds on dual function of regularized RL objective. Another work that shares slight similarity with our method is ReF-ER~\citep{novati2019remember} where they ignore the updates from experiences that deviates significantly from the current policy. Our work focuses on penalizing the TD errors of the samples that deviate from the current policy which leads to smaller priority instead of completely ignoring those experiences. In addition, our proposed method forms a new pipeline of TD-error-based prioritization scheme. 

\section{Conclusion}
By leveraging the regularized RL objective and its dual function, we propose a new pipeline of TD-error-based prioritization scheme that is more robust towards distribution shift between off-policy data and current policy. By considering KL-divergence as the reuglarizer, we formulated a new prioritized experience replay, namely regularized optimal experience replay (ROER). Our proposed ROER when applied to SAC empirically demonstrates the ability of mitigating the underestimation bias and shows faster convergence to the true value. It outperforms baselines in 6 out of 11 continuous control tasks in the online setting and significantly improves the performance in Antmaze with pretraining. However, we recognize the tuning of additional hyperparameters can limit the application. Future work can explore an adaptive loss temperature to dynamically adjust the strength of the regularization. Additionally, it would be valuable to extend the application of the proposed method to offline setting and further explore the applicability to offline-to-online fine tuning. 

\textbf{Limitations.} Although our method provides theoretical motivations and empirically shows performance improvement and convergence to the optimal solutions over various environments, we lack theoretical guarantees to ensure the convergence. Few works have provided the theoretical ground of analyzing convergence using $f$-divergence regularizer~\citep{paternain2019constrained}. More exploration is required to understand the convergence in online setting. 

\subsubsection*{Acknowledgments}
\label{sec:ack}
This research was supported in part by Quanta Computer Inc. and ONR MURI under Grant Number N00014-22-1-2740. We acknowledge access to the LUMI supercomputer, owned by the EuroHPC Joint Undertaking, hosted by CSC (Finland) and the LUMI consortium. J.~Pajarinen was partly supported by Research Council of Finland (345521,353198).

\bibliography{main}
\bibliographystyle{rlc}

\appendix

\section{Derivation Details}
\label{Apdix: proof}
In this section, we provide the detailed derivation of our method for completeness. We reference ~\citet{Nachum2019AlgaeDICEPG} for the derivation. We start by using the regularized max-return objective with divergence term between the on-policy optiaml distribution $d^*$ and off-policy distribution $d^{\mathcal{D}}$
\begin{equation}
    \text{max}_{\pi}\mathcal{J}_{D,f}(\pi):= \mathbb{E}_{(s,a)\sim d^*}[r(s,a)]-\beta D_{f}(d^*||d^{\mathcal{D}}),
\end{equation}
where $\beta > 0$ and $D_f$ denotes the $f$-divergence induced by a convex function $f$:
\begin{align}
    D_f(d^*||d^\mathcal{D}) = \mathbb{E}_{(s,a)\sim d^\mathcal{D}}[f(w_{*/\mathcal{D}}(s,a))],
\end{align}
where $w_{*/\mathcal{D}}:= \frac{d^*(s,a)}{d^\mathcal{D}(s,a)}$.

We then transform the $f$-divergence to its variational form using a dual function $x: S \times A \rightarrow \mathbb{R}$ that is bounded which gives the following expressions 
\begin{align}\label{eq: dual form}
    \tilde{\mathcal{J}}_{D,f}(\pi, x):&=\min_{x}\mathbb{E}_{(s,a)\sim d^*}[r(s,a)] + \beta \cdot \mathbb{E}_{(s,a)\sim d^{\mathcal{D}}}[f_*(x(s,a))] - \beta \cdot \mathbb{E}_{(s,a)\sim d^*}[x(s,a)]\nonumber\\
    &=\min_{x}\mathbb{E}_{(s,a)\sim d^*}[r(s,a)-\beta \cdot x(s,a)] + \beta \cdot \mathbb{E}_{(s,a)\sim d^{\mathcal{D}}}[f_*(x(s,a)].
\end{align}
Here, $f_*$ is the convex conjugate of $f$. Recall the definition of convex conjugate: the convex conjugate of $f(x)$ is defined as $f_*(x) = \text{sup}_{x\in \text{dom}f}\{\langle y, x \rangle - f(x)\}$, where $\langle y, x \rangle$ denotes the dot product~\citep{boyd2004convex}.

To eliminate the dependence on $d^*$, we use change of variables and let $Q(s,a)-\gamma V^*(s') = -\beta x(s,a) + r(s,a)$. Applying the change of variable to Eq.\ref{eq: dual form}, we obtain:
\begin{align}
    \mathcal{J}_{D,f}(\pi, Q):=\min_{Q} \mathbb{E}_{(s,a)\sim d^*}&[r(s,a)+Q(s,a)-\gamma V^*(s')-r(s,a)]\nonumber\\
    &+ \beta \cdot \mathbb{E}_{(s,a)\sim d^{\mathcal{D}}}[f_*(\gamma V^*(s')-Q(s,a) + r(s,a))/\beta].
\end{align}
Note that $B^*Q(s,a)=r(s,a) + \gamma V^*(s')$. We simplify the above as 
\begin{equation}
    \mathcal{J}_{D,f}(\pi, Q):=\min_{Q} \mathbb{E}_{(s,a)\sim d^*}[Q(s,a)-\gamma V^*(s')] + \beta \cdot \mathbb{E}_{(s,a)\sim d^{\mathcal{D}}}[f_*(B^*Q(s,a)-Q(s,a))/\beta].
\end{equation}
Since $x(s,a)$ is bounded and $\gamma<1$, $Q(s,a)$ is also bounded. Define
\begin{equation}
    \Omega_t(s) := \text{Pr}(s=s_t | s\sim \Omega, a_k\sim \pi^*(s_k), s_{k+1}\sim P(\cdot|s_k, a_k) \text{ for } 0\leq k \leq t),
\end{equation}
as the state visitation probability at step $t$ following policy $\pi^*$~\citep{nachum2019dualdice}. Then by telescoping, we have the following process for $\mathbb{E}_{(s,a)\sim d^*}[Q(s,a) - \gamma V^*(s')]$ (denoting this term as $*$)
\begin{align}
    * &= \mathbb{E}_{(s,a)\sim d^*}[Q(s,a) - \gamma \mathbb{E}_{s'\sim P(\cdot|s,a), a'\sim \pi^*(s')}[Q(s', a')]]\nonumber \\
    &= (1-\gamma)\sum^{\infty}_{t=0} \gamma^t\mathbb{E}_{s\sim \Omega_t, a\sim \pi^*(s)} [Q(s,a) - \gamma \mathbb{E}_{s'\sim P(\cdot|s,a), a'\sim \pi^*(s')}[Q(s', a')]]\nonumber\\
    &=(1-\gamma)\sum^{\infty}_{t=1}\gamma^t\mathbb{E}_{s\sim \Omega_t, a\sim \pi^*(s)}[Q(s,a)] - (1-\gamma)\sum^{\infty}_{t=1}\gamma^{t+1}\mathbb{E}_{s\sim\Omega_t, a\sim \pi^*(s)}[Q(s,a)]\nonumber\\
    &=(1-\gamma)\mathbb{E}_{s\sim\Omega, a\sim \pi^*(s)}[Q(s,a)].
\end{align}
Applying the above result to the dual objective, we obtain the final objective:
\begin{equation}
    \mathcal{J}_{D,f}(\pi, Q):=\min_{Q} \beta \cdot \mathbb{E}_{(s,a)\sim d^{\mathcal{D}}}[f_*(B^*Q(s,a)-Q(s,a))/\beta] + (1-\gamma)\mathbb{E}_{s_0\sim \mu_0, a_0 \sim \pi^*(s_0)}[Q(s_0, a_0)]
\end{equation}
which completes the derivation.

\section{Other divergence}
\label{Apdix: other divergence}
In this section, we firstly show the connection between the dual objective and the actor-critc objective. Then we give another consideration of regularizer which results in a different form of prioritization. 

\subsection{Derivation Details of ROER}
KL-divergence has the form $f(x) = x\log(x)$ and its convex conjugate has the form $f_*(y) =  e^{y} - 1$. Let $y=(\mathcal{B}^* Q(s, a) - Q(s, a))/\beta$, we follow the derivation in section~\ref{methods} and obtain the following dual objective:
\begin{equation}
    \min_Q \ \mathbb{E}_{(s, a) \sim d^\mathcal{D}}\left[e^{\left(\mathcal{B}^* Q(s, a) - Q(s, a)\right)/\beta}\right] + (1-\gamma)  \mathbb{E}_{s_0 \sim \mu_0}\left[V^*\left(s_0\right)\right]-1
\end{equation}
which can be expanded to:
\begin{equation}
    \min_Q \ \mathbb{E}_{(s, a) \sim d^\mathcal{D}}\left[e^{\left(\mathcal{B}^* Q(s, a) - Q(s, a)\right)/\beta}\right] + \mathbb{E}_{(s, a, s') \sim d^\mathcal{D}}\left[V^*(s)- \gamma V^*(s')\right]-1.
\end{equation}
Recall that $\gamma V^*(s') = \mathcal{B}^*Q(s,a)-r(s,a) $. We substitute the expression of $\gamma V^*(s')$ back to the above objective and obtain 
\begin{equation}
    \min_Q \ \mathbb{E}_{(s, a) \sim d^\mathcal{D}}\left[e^{\left(\mathcal{B}^* Q(s, a) - Q(s, a)\right)/\beta}\right] + \mathbb{E}_{(s, a, s') \sim d^\mathcal{D}}\left[V^*(s)- \mathcal{B}^*Q(s,a) + r(s,a)\right]-1,
\end{equation}
and we can further simplify the expression and obtain 
\begin{equation}
    \min_Q \ \mathbb{E}_{(s, a) \sim d^\mathcal{D}}\left[e^{\left(\mathcal{B}^* Q(s, a) - Q(s, a)\right)/\beta}\right] - \mathbb{E}_{(s, a, s') \sim d^\mathcal{D}}\left[\mathcal{B}^*Q(s,a) - Q(s,a)\right]-1
\end{equation}
which completes the derivation. This objective corresponds to ExtremeQ loss as in \citet{garg2023extreme}. It has a corresponding ExtremeV loss in the following form:
\begin{equation}
    \mathcal{L}(\mathcal{V}) = \mathbb{E}_{(s, a) \sim d^\mathcal{D}}\left[e^{\left(\mathcal{B}^* Q(s, a) - V(s)\right)/\beta}\right] - \mathbb{E}_{(s, a, s') \sim d^\mathcal{D}}\left[ Q(s,a) - V(s)\right]-1
\end{equation}
which is the objective of our value network.

\subsection{Connection to Actor Critic}
Recall that the dual function of the regularized RL objective with the change of variable has the following form
\begin{equation}
    \mathcal{J}_{D, f}(\pi, Q) = \min_Q \ \beta \cdot \mathbb{E}_{(s, a) \sim d^\mathcal{D}}\left[f_*\left((\mathcal{B}^* Q(s, a) - Q(s, a))/\beta\right)\right] + (1-\gamma)  \mathbb{E}_{s_0 \sim \mu_0, a_0 \sim \pi^*(s_0)}\left[Q\left(s_0, a_0\right)\right]    
\end{equation}
We consider a convex function of the form $f(x)=\frac{1}{2}x^2$. Its convex conjugate has the same form as itself $f_*(y) = \frac{1}{2}y^2$. Let $y:=\mathcal{B}^*Q(s,a)-Q(s,a)$. The above function can be expressed as below:
\begin{equation}
    \mathcal{J}_{D, f}(\pi, Q) = \min_Q \ \frac{1}{2\beta} \cdot \mathbb{E}_{(s, a) \sim d^D}\left[\left(\mathcal{B}^* Q(s, a) - Q(s, a)\right)^2\right] + (1-\gamma)  \mathbb{E}_{s_0 \sim \mu_0, a_0 \sim \pi^*(s_0)}\left[Q\left(s_0, a_0\right)\right]
\end{equation}
 which transforms the off-policy actor-critic to an on-policy actor-critic by introducing the second term. This unifies the two separate objectives of value and policy into a single objective and both functions are trained with respect to the same off-policy objective~\citep{Nachum2019AlgaeDICEPG}.
 
\subsection{Pearson $\chi^2$ Divergence}
A variety of $f$-divergence can be suitable candidates for the dual objective and prioritization derivation. Here, we provide a list of $f$-divergences $f(x)$, its corresponding convex conjugates $f_*(y)$ and the potential priority forms $f_*'(y)$ in Table~\ref{tab: divergence}. We note that the forms presented are theoretical forms and they may vary when applying to the RL objectives. 

\begin{table}[h]
\begin{center}
\def\arraystretch{1.5}
\begin{tabular}{llll}
\hline
Divergence & $f(x)$ & $f_*(y)$ & $f_*'(y)$ \\ \hline
KL & $x\log x$ & $e^y-1$ & $d^y$\\
Reverse KL & $-\log x$ & $-\log(1-y)$ & $\frac{1}{1-y}$\\
Pearson $\chi^2$ & $\frac{1}{2}(x-1)^2$ & $\frac{1}{2}y^2 + y$ & $y+1$ \\
Neyman $\chi^2$ & $\frac{(x-1)^2}{2x}$ & $-\sqrt{1-2y}+1$ & $\frac{1}{\sqrt{1-2y}}$ \\
Total variation & $\frac{1}{2}|x-1|$ & $y$ & $1$\\
Squared Hellinger & $2(\sqrt{x}-1)^2$ & $\frac{2y}{2-y}$ & $\frac{4}{(2-y)^2}$ \\
\hline
\end{tabular}
\end{center}
\caption{List of $f$-divergence functions $f(x)$, convex conjugates $f_*(y)$ and the potential priority forms $f_*'(y)$.}
\label{tab: divergence}
\end{table}

We use Pearson $\chi^2$ divergence as an example as the resulting objective has a particular implication. Pearson $\chi^2$ divergence has the form $f(x) = \frac{1}{2}(x-1)^2$ and its convex conjugate has the form $f_*(y) = \frac{1}{2}y^2 + y$. Again, let $y:=\mathcal{B}^*Q(s,a)-Q(s,a)$. We can obtain the following dual objective:
\begin{equation}
    \min_Q \ \frac{1}{2\beta} \cdot \mathbb{E}_{(s, a) \sim d^D}\left[\left(\mathcal{B}^* Q(s, a) - Q(s, a)\right)^2\right] + \mathbb{E}_{(s, a) \sim d^D}\left[\mathcal{B}^* Q(s, a) - Q(s, a)\right] + (1-\gamma)  \mathbb{E}_{s_0 \sim \mu_0}\left[V^*\left(s_0\right)\right]
\end{equation}
Using $\gamma V^*(s') + r(s,a) = \mathcal{B}^*Q(s,a)$, the objective can be further simplified to:
\begin{equation}
    \min_Q \ \frac{1}{2\beta} \cdot \mathbb{E}_{(s, a) \sim d^D}\left[\left(\mathcal{B}^* Q(s, a) - Q(s, a)\right)^2\right] + \mathbb{E}_{(s, a) \sim d^D}\left[V^*(s)- Q(s, a)\right]
\end{equation}
We note that this corresponds to the learning objective of conservative Q-learning~\citep{kumar2020conservative}. This objective is optimized when $d^*/d^D = f_*'(\delta_Q^*/\beta) = \delta_Q/\beta + 1$ which gives a new form of priority calculation. We address that even though this form is almost identical to PER, the source of $\delta$ is different. We derive this priority form from the conservative Q-learning objective and thus, it requires the value network to use the corresponding loss function. In addition, the loss temperature $\beta$ here also controls the scale of the TD error and the strength of the regularizer which we expect to give better performance than the naive PER.




\section{Experiments}
\label{Apdix: experiments}
In this section, we provide details of the implementation and the experiments with further discussion on hyperparameter ablation and selection. 

\subsection{Experimental Details}

\textbf{Environment} 

In the online setting, our agents are evaluated in MuJoCo via OpenAI gym interface using the v2 environments~\citep{brockman2016openai} and DM Control tasks~\citep{tunyasuvunakool2020dm_control}. For the MuJoCo environment, we do not modify or preprocess the state space, action space and reward function for easy reproducibility. For the DM Control tasks, we adapt to the gym environment interface. In the online with pretraining settings, our agents are evaluated in the environment with D4RL datasets~\citep{fu2020d4rl}. We shaped the reward of Antmaze by subtracting 1 as suggested in \citet{kumar2020conservative} which shows to largely benefit the performance in Antmaze. Each environment runs for a maximum of 1000 time steps which is the default setting or until a termination state is reached. 

\textbf{Value estimation} 

Value estimates are averaged over mini-batches of 256 and sampled every 2000 iterations. The true value is estimated by sample 256 state-action pairs from the replay buffer and compute the discounted return by running the episode following the current policy until termination.

\textbf{Reward Evaluation} 

In the online setting, we evaluate the current policy over 10 episodes for every 5000 training steps. The evaluation reward takes the average over the 10 episodes. In the online with pretraining setting, we evaluate the current policy over 100 episodes for every 10000 training steps due to the difficulty of the environments. The evaluation reward takes the average over the 100 episodes. 

\textbf{Algorithm implementation} 

We base our implementation of SAC off \citet{jaxrl}. It uses one target critic, double critics, a single actor network and a single network for temperature adjustment for maximum entropy. We add an additional value network with extreme q-learning loss for ROER TD error estimation and priority calculation. We use the default hyperparameters and network architectures for the SAC algorithms for all of our experiments. The hyperparameters and the network architecture are shown in Table~\ref{tab: network arch}.

\begin{table}[h]
\begin{center}
\begin{tabular}{ll}
\hline
Parameter & Value \\ \hline
optimizer & Adam\\
learning rate & $3\times 10^{-3}$\\
actor, critic, and value network arch & (256, 256)\\
non-linearity & ReLU\\
value network noise & 0.1\\
batch size & 256\\
buffer size & 1,000,000 , 2,000,000(antmaze)\\
discount & 0.99\\
target smoothing coefficient & $5 \times 10^{-3}$\\
gradient penalty coefficient & 1\\
\hline
\end{tabular}
\end{center}
\caption{Hyperparameters of SAC and network architecture.}
\label{tab: network arch}
\end{table}

We adapt the code in \citet{lahire2021large} into JAX implementation for LaBER. We adapt the proposed method in \citet{fujimoto2020equivalence} for PER implementation and uses the loss adjusted version of PER as it shows similar or even better performance than the original implementation. Loss adjusted PER (LAP) uses Huber loss for critic objective which has the following form~\citep{huber1992robust}
\begin{equation}
  \mathcal{L}_{\text{Huber}}(\delta(i)) =
    \begin{cases}
      0.5 \delta(i)^2 & \text{if $|\delta(i)|\leq k $}\\
      k(|\delta(i)|-0.5k) & \text{otherwise}\\
    \end{cases}       
\end{equation}
where $k$ is the bound for the Huber loss transformation and the default value is 1. We also uses Huber loss for our proposed ROER in the same idea as LAP. For SAC with uniform experience replay, we keep the original loss form which uses mean square loss. To stabilize the performance of all algorithms used in this study, we apply an additional gradient penalty to the critic loss inspired by~\citet{petzka2017regularization} which penalizes gradient with norm great than 1. The penalty $\kappa$ has the following form
\begin{equation}
    \kappa = \text{max}(\|\nabla Q_\theta -1\|, 0)^2
\end{equation}

We note ROER uses exponential function in the formulation and it is sensitive to outliers. Thus, we lower clip the immediate weight with value 1 and use batch mean normalization on the immediate weight $e^{\delta_Q/\beta}$ as following
\begin{equation}
    e^{\delta_Q/\beta}_{\text{normalized}} = \frac{e^{\delta_Q/\beta}}{\bar{e}^{\delta_Q/\beta}}
\end{equation}
where $\bar{e}^{\delta_Q/\beta}$ denotes the batch mean. To further stabilize the performance and prevent the outliers, we clip the immediate weight and add minimum clip on the final priority which are further discussed in the following section of ablation study and hyperparameter selections.

\subsection{Hyper-parameter Selection}
\textbf{Online} 

For the value of parameter of PER, we use \citet{fujimoto2020equivalence} as a reference for MuJoCo environment where they show that a weight scale $\alpha=0.4$ works best for the set of tasks. As to DM Control, we search over the set $[0.1, 0.2, 0.4, 0.6, 0.8]$ for individual task. The final choice of the value for each task is shown in Table~\ref{tab: hyperparameter online}.

For the value of parameter large batch of LaBER, we search over the set $[768, 1024, 1280, 1536]$ for individual task in both MuJoCo and DM Control environment. The final choice of the value for each task is shown in Table~\ref{tab: hyperparameter online}.

\begin{figure}[t!] 
\begin{subfigure}{0.19\textwidth}
\includegraphics[width=\linewidth]{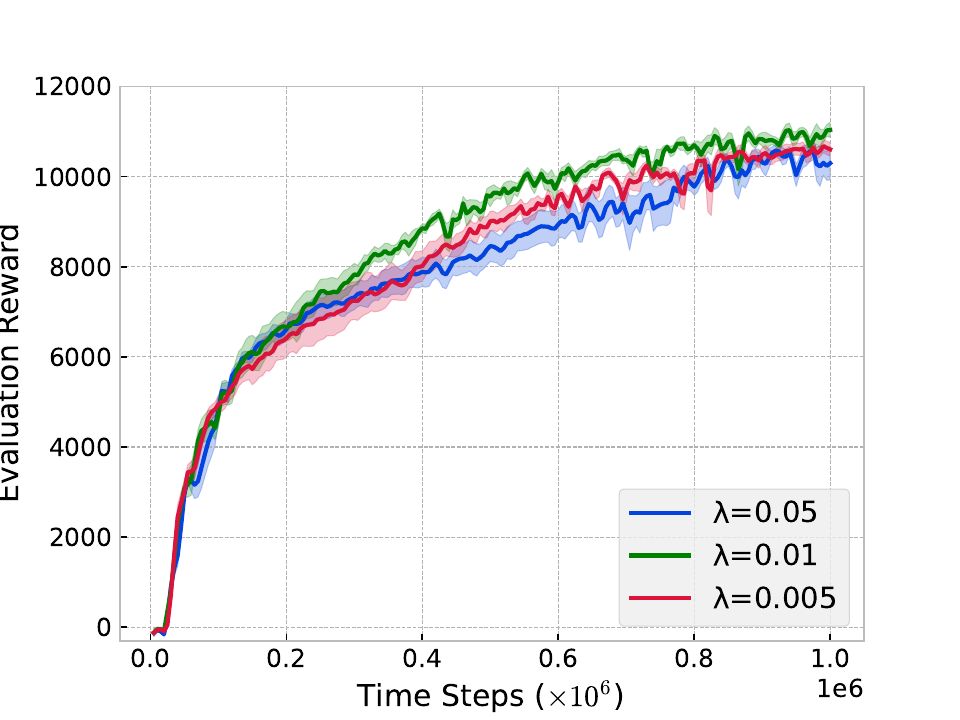}
\caption{$\lambda$} \label{fig:ablation halfcheetah convergence rate}
\end{subfigure}
\begin{subfigure}{0.19\textwidth}
\includegraphics[width=\linewidth]{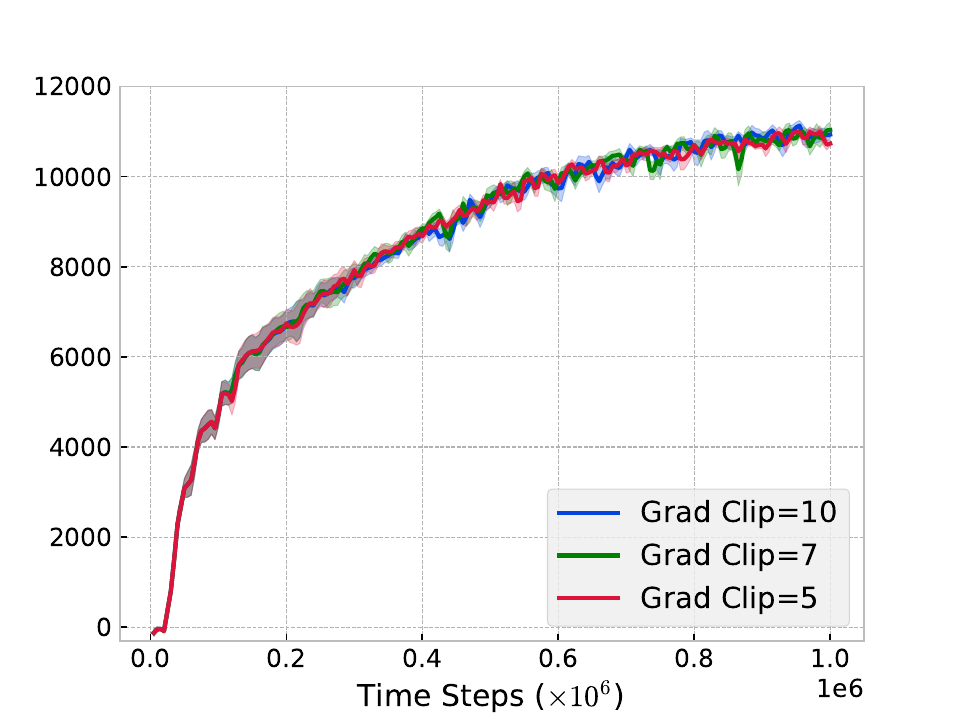}
\caption{Grad Clip} \label{fig:ablation halfcheetah grad clip}
\end{subfigure}
\begin{subfigure}{0.19\textwidth}
\includegraphics[width=\linewidth]{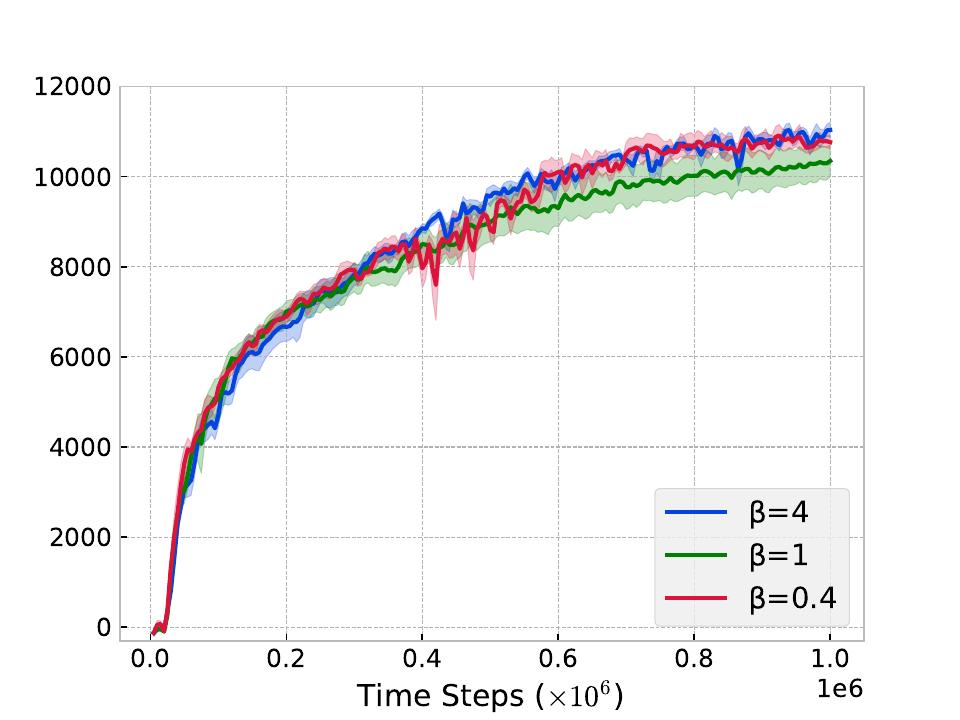}
\caption{$\beta$} \label{fig:ablation halfcheetah loss temp}
\end{subfigure}
\begin{subfigure}{0.19\textwidth}
\includegraphics[width=\linewidth]{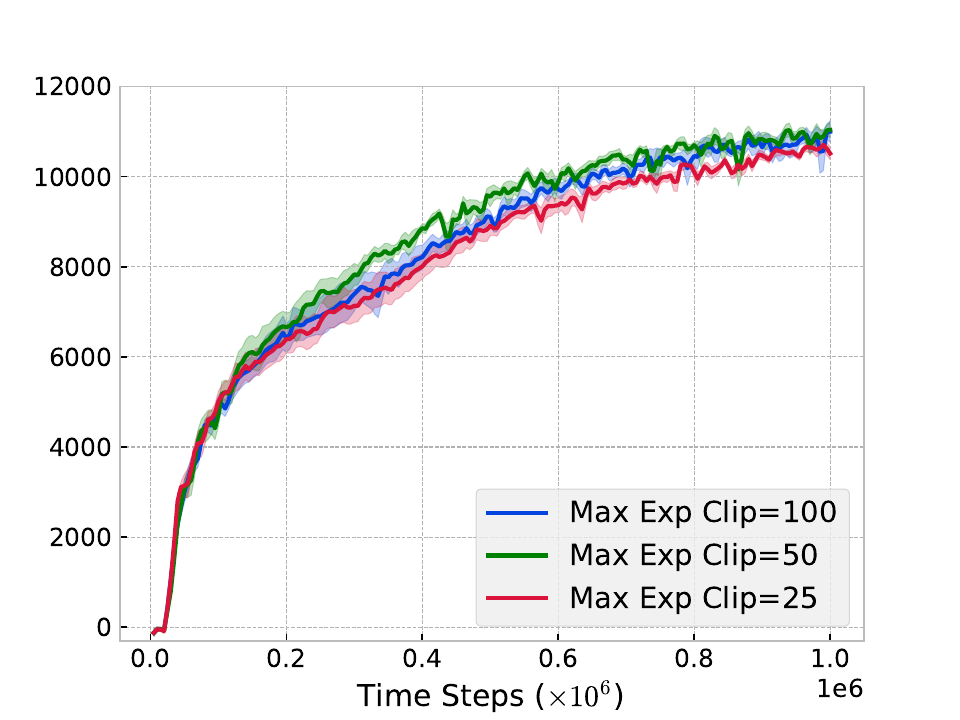}
\caption{Max Exp Clip} \label{fig:ablation halfcheetah max exp clip}
\end{subfigure}
\begin{subfigure}{0.19\textwidth}
\includegraphics[width=\linewidth]{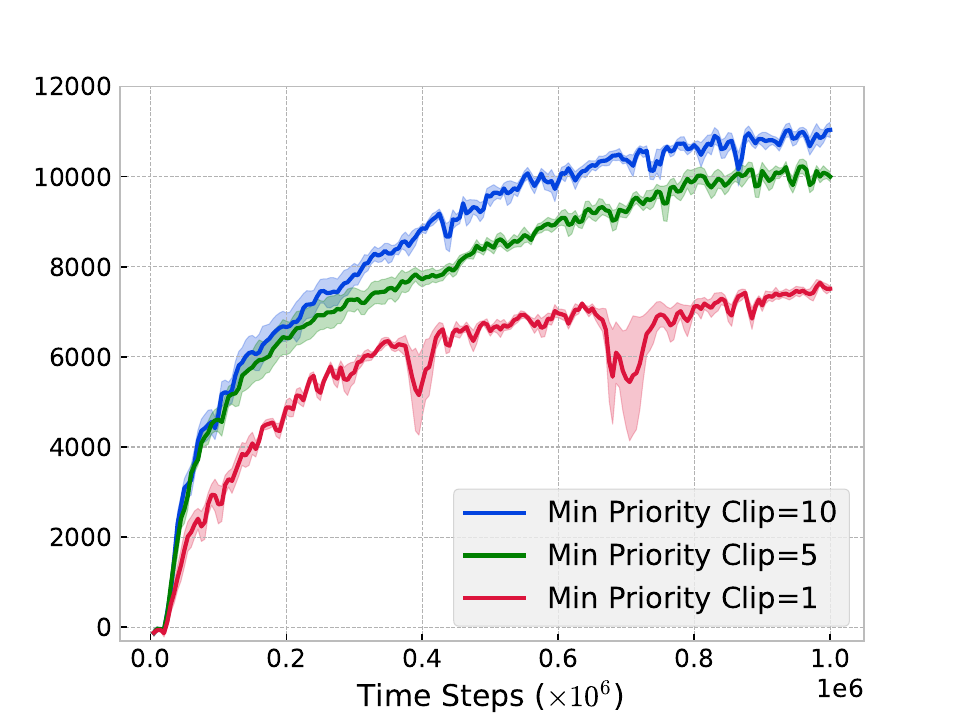}
\caption{Min Clip} \label{fig:ablation halfcheetah min priority clip}
\end{subfigure}
\caption{Convergence rate ($\lambda$), Gumbel loss clip (Grad Clip), loss temperature ($\beta$), Maximum exponential clip (Max Exp Clip), and minimum priority clip (Min Clip) Ablation for HalfCheetah-v2 over 5 random seeds. One parameter is changing while the rest are fixed. The default combination is [0.01, 7, 4, 50, 10] which is the set used in our final results. All curves are smoothed with Savitzky–Golay filter for visual clarity. The shaded region represents standard error which is favored in this case to separate the curves.} \label{fig:ablation halfcheetah}
\end{figure}

\begin{figure}[t!] 
\begin{subfigure}{0.19\textwidth}
\includegraphics[width=\linewidth]{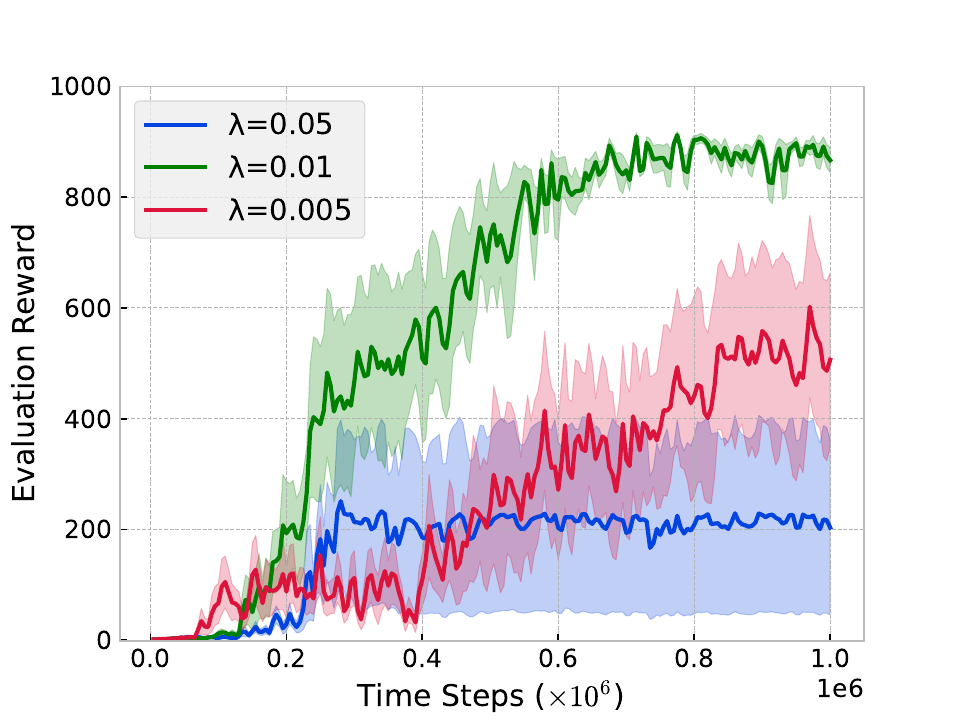}
\caption{$\lambda$} \label{fig:ablation hopperstand convergence rate}
\end{subfigure}
\begin{subfigure}{0.19\textwidth}
\includegraphics[width=\linewidth]{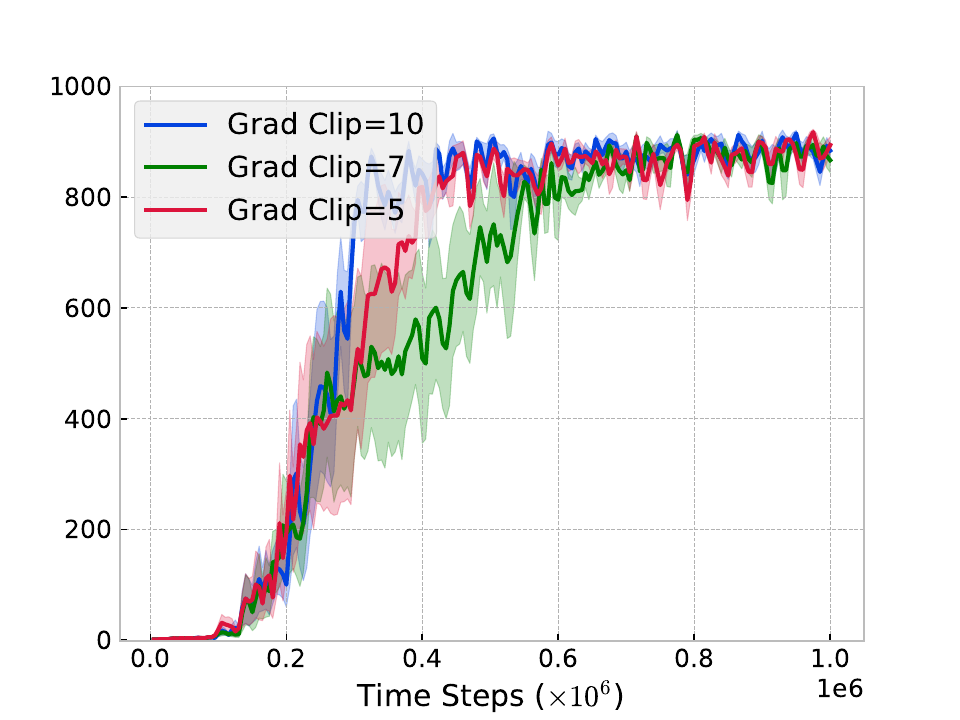}
\caption{Grad Clip} \label{fig:ablation hopperstand grad clip}
\end{subfigure}
\begin{subfigure}{0.19\textwidth}
\includegraphics[width=\linewidth]{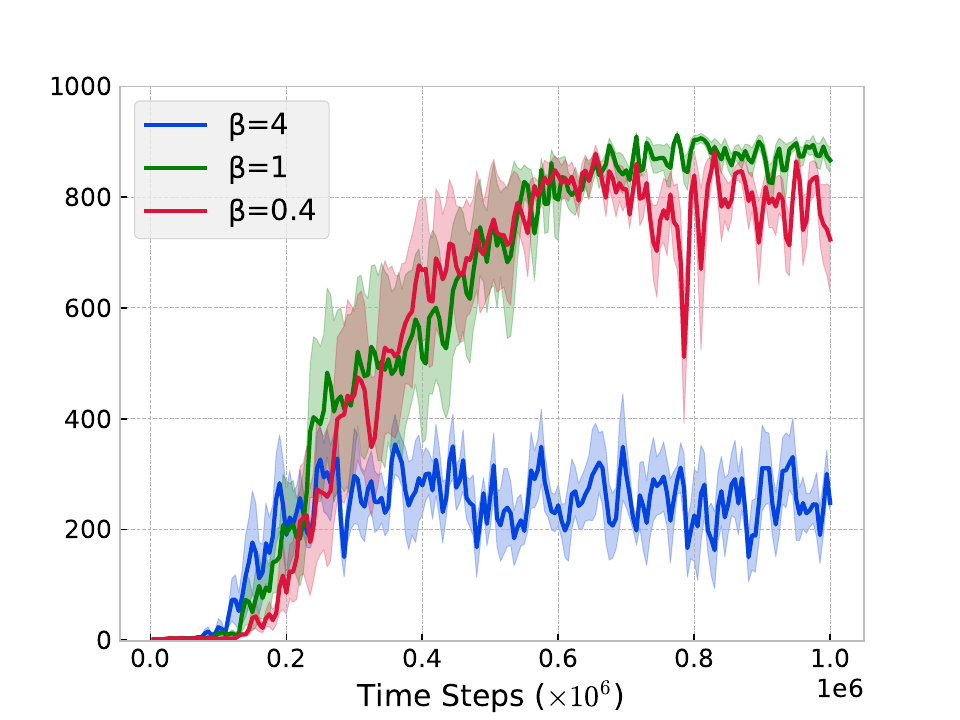}
\caption{$\beta$} \label{fig:ablation hopperstand loss temp}
\end{subfigure}
\begin{subfigure}{0.19\textwidth}
\includegraphics[width=\linewidth]{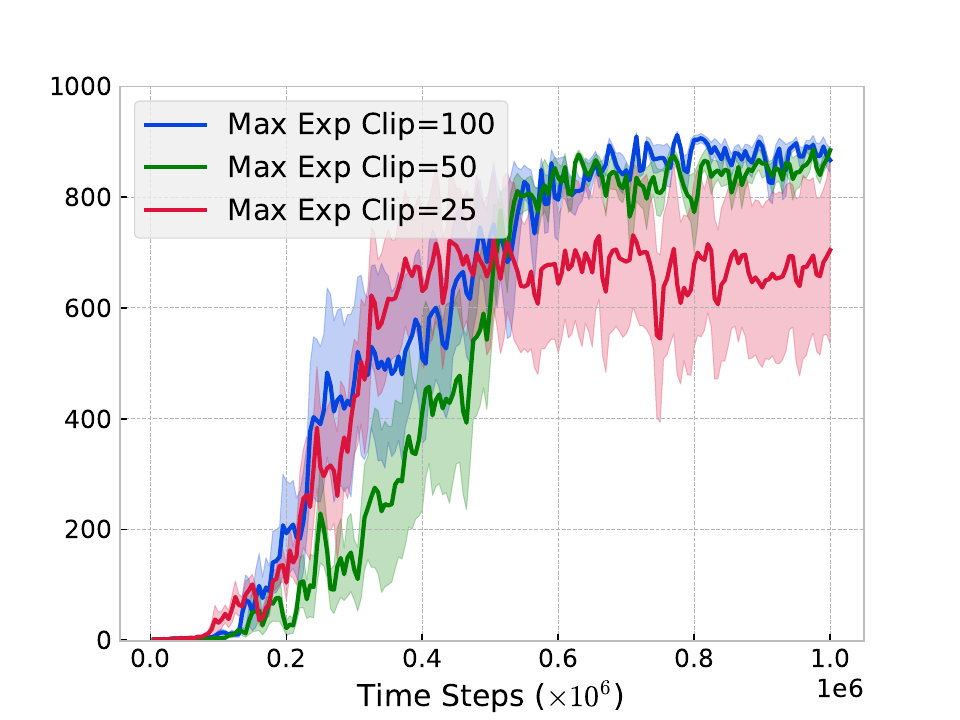}
\caption{Max Exp Clip} \label{fig:ablation hopperstand max exp clip}
\end{subfigure}
\begin{subfigure}{0.19\textwidth}
\includegraphics[width=\linewidth]{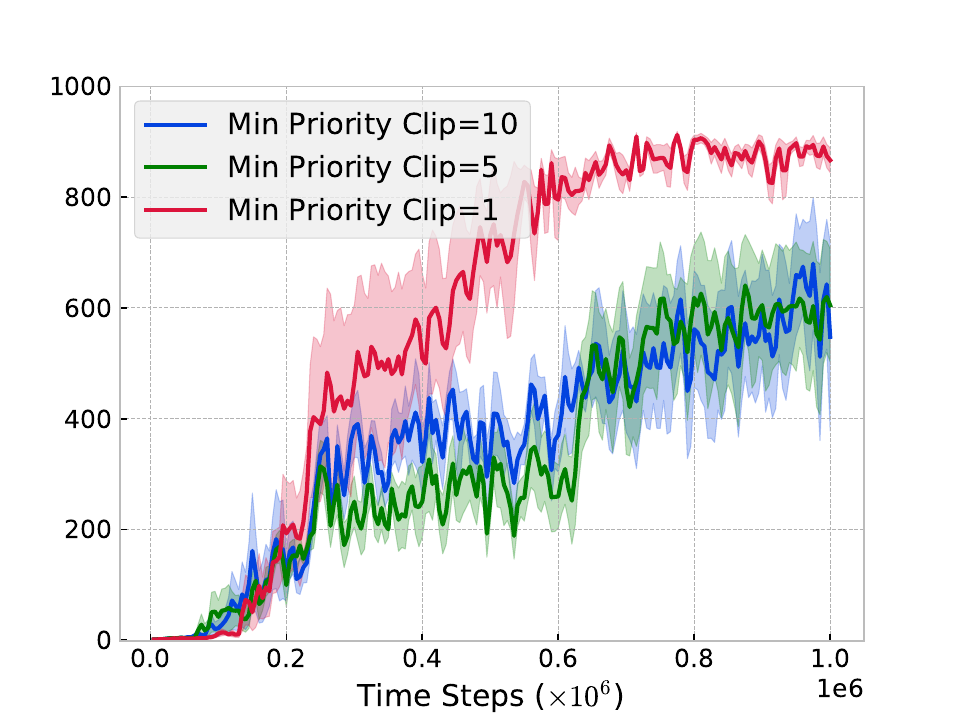}
\caption{Min Clip} \label{fig:ablation hopperstand min priority clip}
\end{subfigure}
\caption{Convergence rate ($\lambda$), Gumbel loss clip (Grad Clip), loss temperature ($\beta$), Maximum exponential clip (Max Exp Clip), and minimum priority clip (Min Clip) Ablation for Hopper-stand over 5 random seeds. One parameter is changing while the rest are fixed. The default combination is [0.01, 7, 1, 10, 1] which is the set used in our final results. All curves are smoothed with Savitzky–Golay filter for visual clarity. The shaded region represents standard error which is favored in this case to separate the curves.} \label{fig:ablation hopper stand}
\end{figure}

For our proposed method ROER, we have 5 parameters, namely convergence rate ($\lambda$), gumbel loss clip (Grad Clip), loss temperature ($\beta$), immediate weight clip (Max Exp Clip), and minimum priority clip (Min Clip) that require tuning. However, we discover that only $\beta$ and the clip range requires tuning for each specific environment while we can use a set of values for the rest. We use the set of value used in \citet{garg2023extreme} as a reference and search over the set $[0.005, 0.01, 0.05]$ for $\lambda$, $[5, 7, 10]$ for the Grad Clip, $[0.4, 1, 4]$ for $\beta$, $[25, 50, 100]$ for the Max Exp Clip, and $[1, 5, 10]$ for the Min Priority Clip. The final choice of the value for each task is shown in Table~\ref{tab: hyperparameter online}. We take HalfCheetaah-v2 from MuJoCo and Hopper-Stand from DM Control as examples to show the hyper-parammeter ablations for the online experiments. We show the effect of each parameter by fixing the rest as the default values. The default parameter values for HalfCheetah-v2 is $ \lambda=0.01, \text{Grad Clip}=7, \beta=4, \text{Max Exp Clip}=50, \text{Min Clip}=10$. The default parameter values for Hopper-Stand is $ \lambda=0.01, \text{Grad Clip}=7, \beta=1, \text{Max Exp Clip}=100, \text{Min Clip}=1$. The performance comparisons of the two task for varying each parameter are plotted in Fig.~\ref{fig:ablation halfcheetah} and Fig.~\ref{fig:ablation hopper stand} respectively.

\begin{table}[http]
\begin{center}
\resizebox{\columnwidth}{!}{
\begin{tabular}{l|ccccc|cc|c}
\hline
 &  &  & ROER &  & & PER& & LaBER\\
Env & $\lambda$ & $\beta$ & Grad Clip & Min Priority Clip & Max Exp Clip & $\alpha$ & Min Priority Clip & Large Batch\\ \hline
Ant-v2 & 0.01 & 1 & 7 & 10 & 100 & 0.4 & 1 & 1280\\
HalfCheetah-v2 & 0.01 & 4 & 7 & 10 & 50 & 0.4 & 1 & 1024\\
Hopper-v2 & 0.01 & 0.4 & 7 & 10 & 100 & 0.4 & 1 & 1536\\
Humanoid-v2 & 0.01 & 4 & 7 & 10 & 50 & 0.4 & 1 & 768\\
Walker2d-v2 & 0.01 & 4 & 7 & 10 & 50 & 0.4 & 1 & 1024\\ \hline
Fish-swim & 0.01 & 1 & 7 & 1 & 100 & 0.4 & 1 & 768\\
Hopper-hop & 0.01 & 1 & 7 & 1 & 50 & 0.6 & 1 & 1024\\
Hopper-stand & 0.01 & 1 & 7 & 1 & 100 & 0.2 & 1 & 1280\\
Humanoid-run & 0.01 & 4 & 7 & 1 & 100 & 0.4 & 1 & 1536\\
Humanoid-stand & 0.01 & 4 & 7 & 1 & 100 & 0.4 & 1 & 1280\\
Quadruped-run & 0.01 & 1 & 7 & 10 & 100 & 0.4 & 1 & 1024\\\hline
\end{tabular}}
\end{center}
\caption{Hyperparameters for ROER, PER and LaBER in online setting}
\label{tab: hyperparameter online}
\end{table}

According to Fig.~\ref{fig:ablation halfcheetah convergence rate} and Fig.~\ref{fig:ablation hopperstand convergence rate}, $\lambda=0.01$ achieves the best performance for the two environments. We note that a too big $\lambda$ can results in divergence as in Fig.~\ref{fig:ablation hopperstand convergence rate} where $\lambda=0.05$ achieves bad performance due to the too quick priority update which results in numerical instability. A small $\lambda$ gives stable convergence but too small $\lambda$ may slow down the convergence in some cases as in Fig.~\ref{fig:ablation hopperstand convergence rate}. We discover that generally $\beta=0.01$ works well across a wide domain and we use this value for all environments in our study. 

Grad Clip is initially leveraged to prevent the outliers in gumbel loss of extreme q-learning~\citep{garg2023extreme}. In our case, we found that varying its value has negligible effect on the final result. As shown in Fig.~\ref{fig:ablation halfcheetah grad clip} and Fig.~\ref{fig:ablation hopperstand grad clip}, all three values give similar performance. We use 7 as the value for all environments in our study.

The loss temperature $\beta$ controls the strength of the penalization from the KL regularizer on the distributions between the on-policy data and off-policy data in replay buffer. It also scales the TD error and thus affect the value of the priority. Theoretically, a small $\beta$ is beneficial for datasets with lots of random noisy actions and far from the on-policy distributions while a high $\beta$ works well for datasets close to the on-policy distribution. In practice, we note that using a fixed value requires tuning for specific environments due to the different dynamics. $\beta=4$ works well for most tasks in MuJoCo while $\beta=1$ works well for most tasks in DM Control. The affect of $\beta$ is more obvious and easier to interpret in online with pretraining setting which is discussed in the following subsection. We recognize the difficulty of deciding the value for $\beta$ and a solution to this can be using adaptive loss temperature. As the off-policy data in replay buffer converges to the on-policy distribution, we increase the value of $\beta$.

The Max Exp Clip and Min Priority Clip serve to prevent outliers and control the range of priority distribution. In most cases, a Max Exp Clip of value 50 or 100 works well. Min Clip requires tuning for each environment but we find Min Priority Clip $=10$ works well for all MuJoCo tasks and Min Priority Clip $=1$ works well for most of DM Control tasks except quadruped-run which uses value 10. A lower Max Exp Clip and a higher Min Priority Clip reduce the range of distribution and tend to stabilize the performance but may give sub-optimal performance. In addition, a higher Min Priority Clip can also prevent experience forgetting as the downweighted samples do not differ too much from the rest.

\textbf{Online with Pretraining} 

To select the appropriate value of $\alpha$ for PER, we searched over $[0.1, 0.2, 0.4, 0.6, 0.8, 1]$. To select the appropriate value of large batch for LaBER, we search over $[768, 1024, 1280, 1536]$. The final choices for each environment are listed in Table~\ref{tab: hyperparameter antmaze}.

To select the appropriate parameter values for ROER, we searched over the same sets of values for $\lambda$, Grad Clip, Max Exp Clip, and Min Priority Clip as in the online setting. For the loss temperature $\beta$, we consider a slightly larger set $[0.4, 0.8, 2, 4]$ to better demonstrate the effect of $\beta$. Similar to the online setting, we also found that a set of values work well across different environments and the final choices are shown in Table~\ref{tab: hyperparameter antmaze}. We take Antmaze-Umaze with pretraining using antmaze-umaze-diverse-v2 dataset as an example to ablate the hyperparameters. While varying the value of one parameter, we fix the rest as the default values. The default parameter values for antmaze-umaze-diverse-v2 is $ \lambda=0.01, \text{Grad Clip}=7, \beta=0.4, \text{Max Exp Clip}=50, \text{Min Clip}=1$.

\begin{table}[http]
\begin{center}
\resizebox{\columnwidth}{!}{
\begin{tabular}{l|ccccc|cc|c}
\hline
 &  &  & ROER &  & & PER& & LaBER\\
Env & $\lambda$ & $\beta$ & Grad Clip & Min Priority Clip & Max Exp Clip & $\alpha$ & Priority Clip & Large Batch\\ \hline
antmaze-umaze-v2 & 0.01 & 0.4 & 7 & 10 & 50 & 0.1 & 1 & 1024\\
antmaze-umaze-diverse-v2 & 0.01 & 0.4 & 7 & 1 & 50 & 0.4 & 1 & 1024\\
antmaze-mediumm-play-v2 & 0.01 & 0.4 & 7 & 1 & 50 & 0.4 & 1 & 1536\\
antmaze-medium-diverse-v2 & 0.01 & 0.4 & 7 & 1 & 50 & 0.4 & 1 & 1024\\ \hline
\end{tabular}}
\end{center}
\caption{Hyperparameters for ROER, PER, and LaBER in online with pretraining setting}
\label{tab: hyperparameter antmaze}
\end{table}

\begin{figure}[h] 
\begin{subfigure}{0.19\textwidth}
\includegraphics[width=\linewidth]{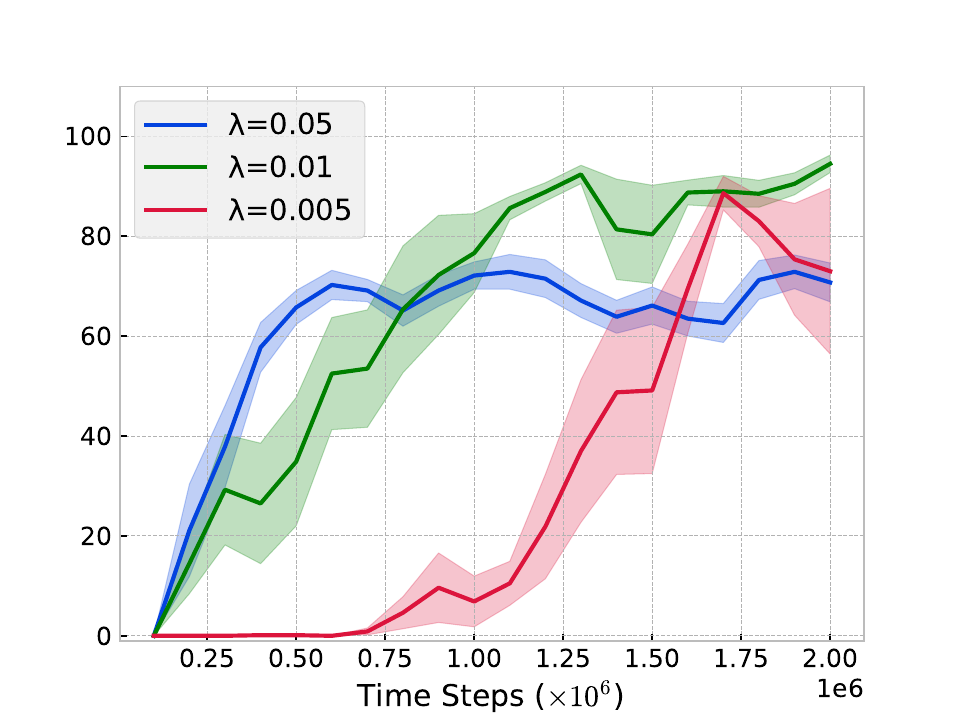}
\caption{$\lambda$} \label{fig:ablation antmaze convergence rate}
\end{subfigure}
\begin{subfigure}{0.19\textwidth}
\includegraphics[width=\linewidth]{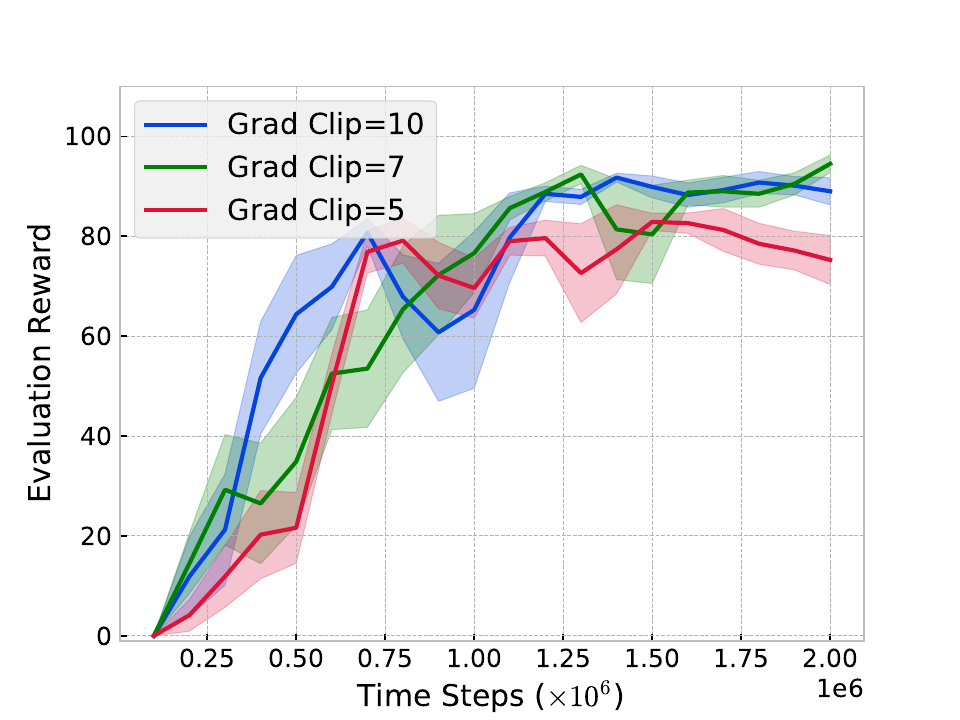}
\caption{Grad Clip} \label{fig:ablation antmaze grad clip}
\end{subfigure}
\begin{subfigure}{0.19\textwidth}
\includegraphics[width=\linewidth]{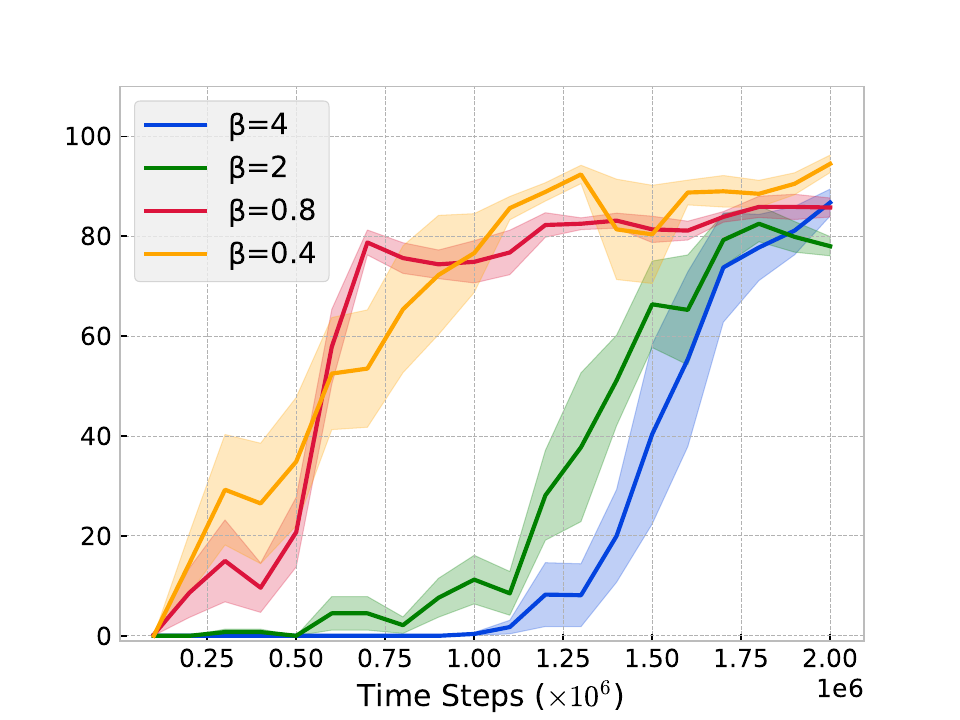}
\caption{$\beta$} \label{fig:ablation antmaze loss temp}
\end{subfigure}
\begin{subfigure}{0.19\textwidth}
\includegraphics[width=\linewidth]{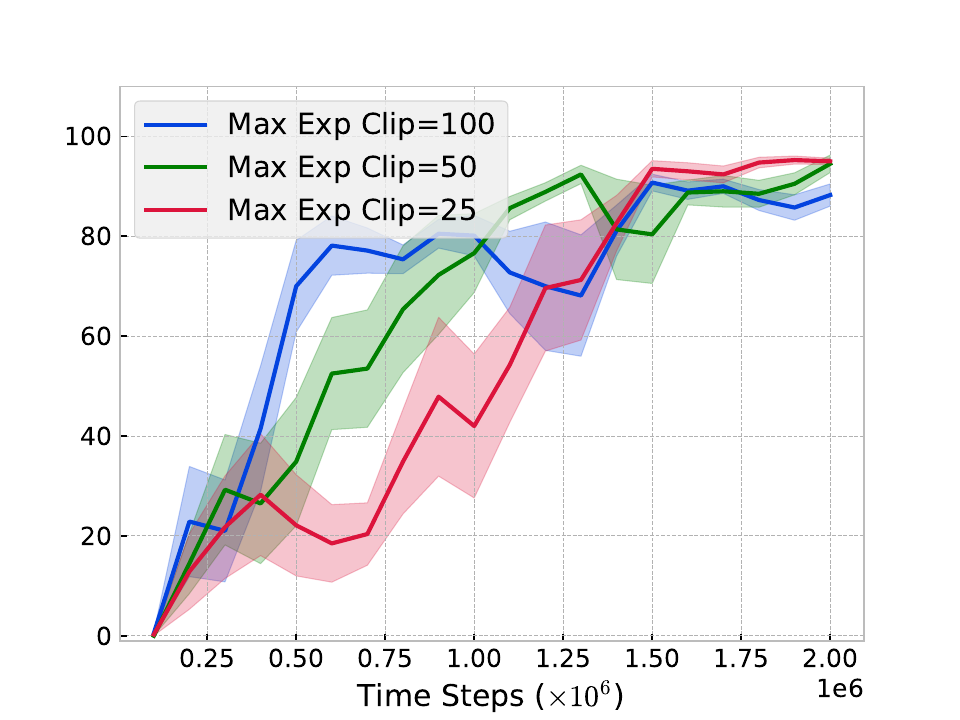}
\caption{Max Exp Clip} \label{fig:ablation antmaze exp clip}
\end{subfigure}
\begin{subfigure}{0.19\textwidth}
\includegraphics[width=\linewidth]{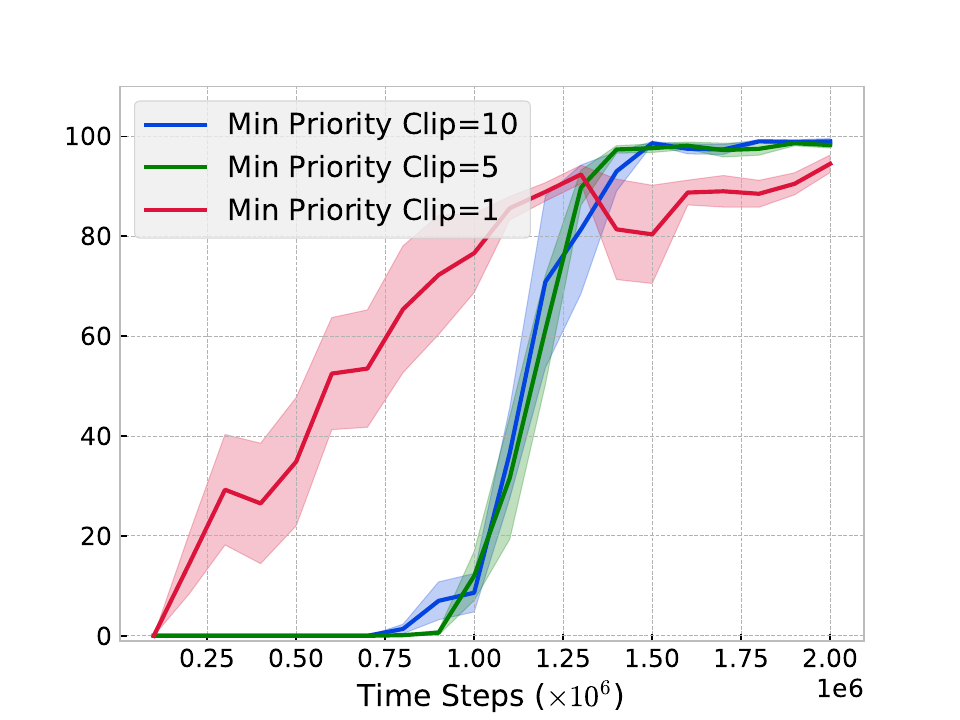}
\caption{Min Clip} \label{fig:ablation antmaze min priority clip}
\end{subfigure}
\caption{Gumbel loss clip (Grad Clip), loss temperature ($\beta$), and minimum priority clip (Min Clip) Ablation for Antmaze with Antmaze-umaze-diverse-v2 dataset over 5 random seeds. One parameter is changing while the rest are fixed. The default combination is [0.01, 7, 0.4, 50, 1] which is the set used in our final results. The shaded region represents standard error which is favored in this case to separate the curves.} \label{fig:ablation antmaze}
\end{figure}

Similar to the online setting, $\lambda=0.01$ shows good performance and the value of Grad Clip only slightly affect the final performance. We choose the values for the two parameters same as before. Varying $\beta$ shows that a smaller value results in early performance improvement while bigger value only shows improvement after the on-policy data overweight in the replay buffer as demonstrated in Fig.~\ref{fig:ablation antmaze loss temp}. This confirms our previous discussion that a small $\beta$ is beneficial for datasets that differ from the on-policy distribution while a high $\beta$ favors datasets close to the on-policy distribution. The behavior of varying Max Exp Clip and Min Priority Clip also corresponds to our previous discussion that a smaller range of priority distribution gives a stable but potentially sub-optimal performance while a wider range can benefit the agent to explore and potentially learn faster.

\section{Additional Results}
\label{Apdix: additional results}
\begin{figure}[h] 
\begin{center}
\begin{subfigure}{0.325\textwidth}
\includegraphics[width=\linewidth]{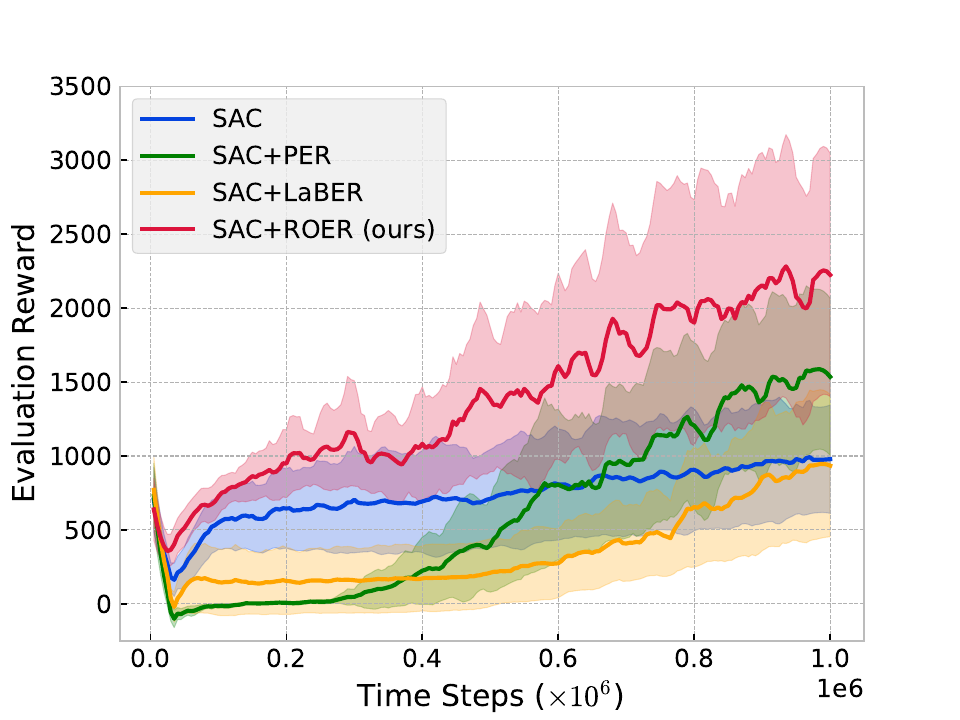}
\caption{Ant-v2}\label{fig:Ant-v2}
\end{subfigure}
\begin{subfigure}{0.325\textwidth}
\includegraphics[width=\linewidth]{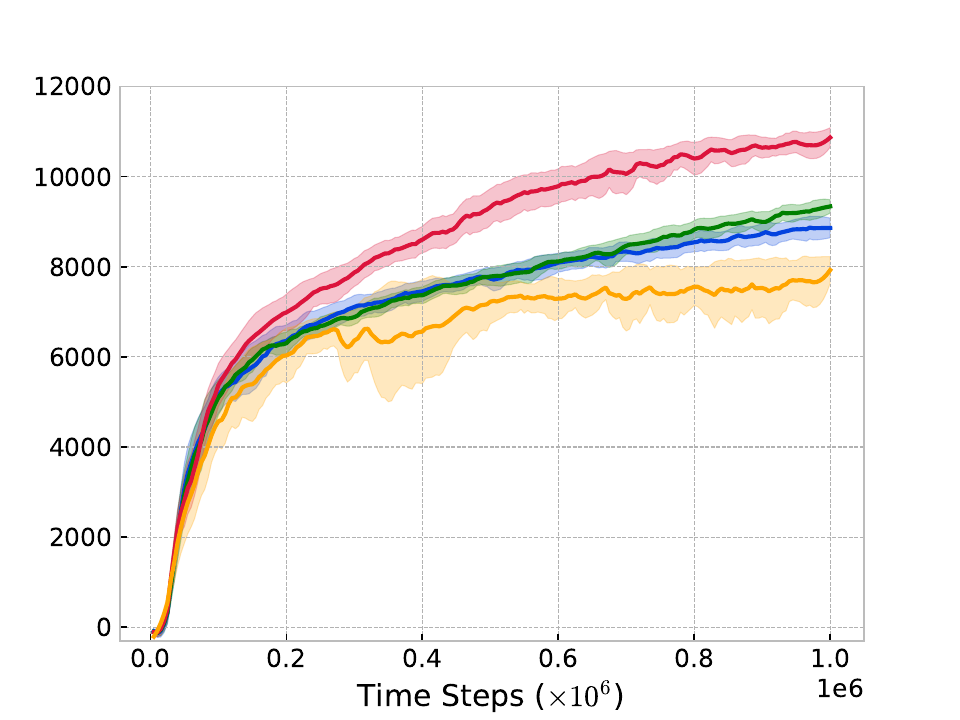}
\caption{HalfCheetah-v2}\label{fig: HalfCheetah-v2}
\end{subfigure}
\begin{subfigure}{0.325\textwidth}
\includegraphics[width=\linewidth]{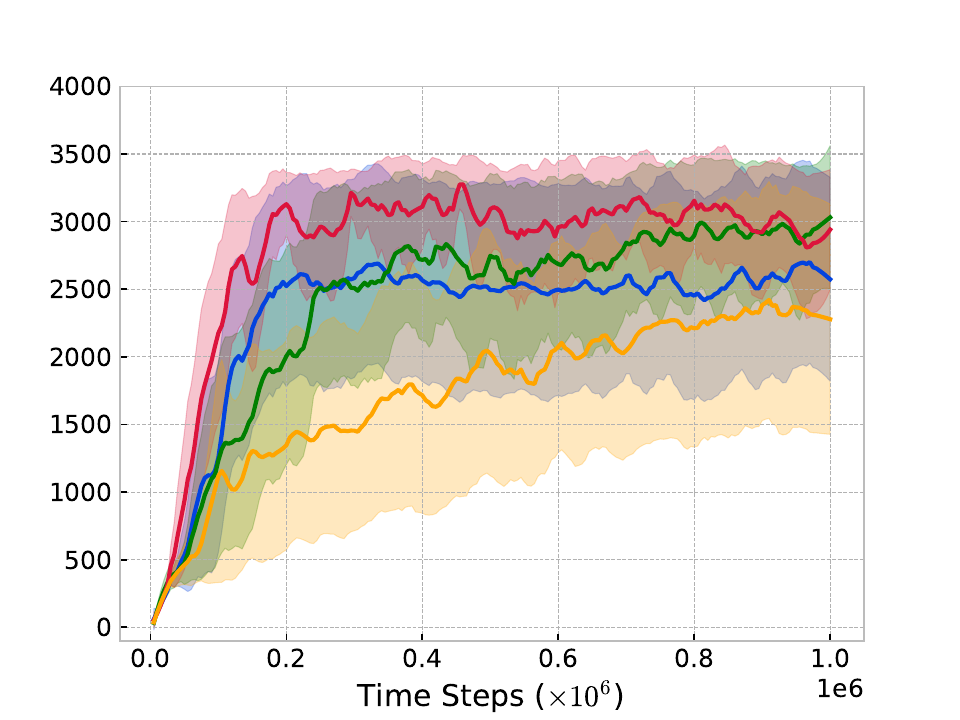}
\caption{Hopper-v2}\label{fig:Hopper-v2}
\end{subfigure}

\begin{subfigure}{0.325\textwidth}
\includegraphics[width=\linewidth]{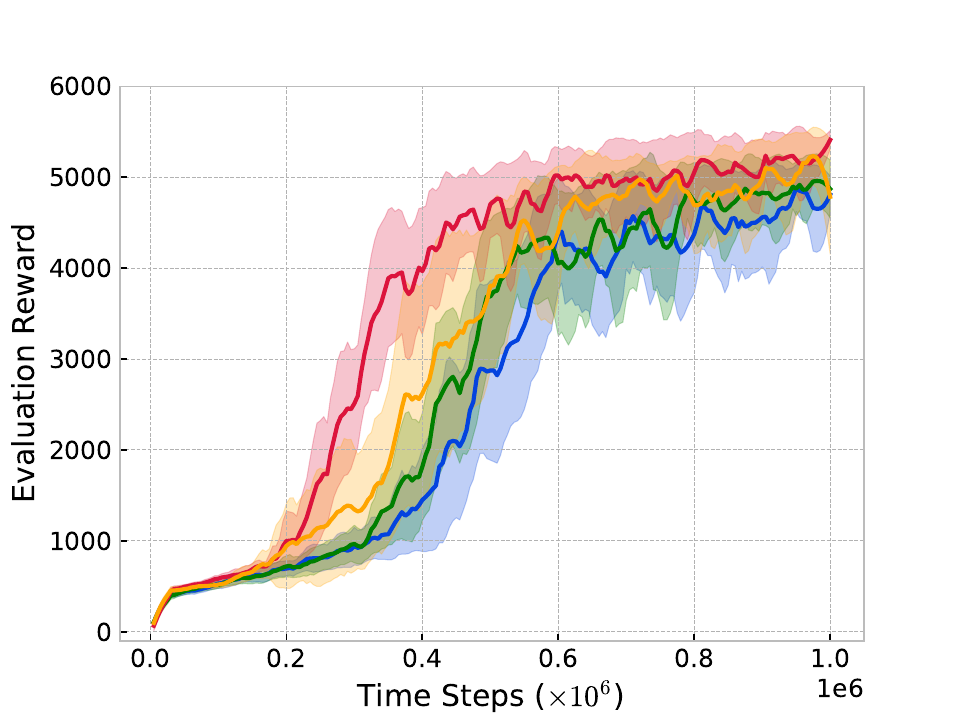}
\caption{Humanoid-v2}\label{fig:Humanoid-v2}
\end{subfigure}
\begin{subfigure}{0.325\textwidth}
\includegraphics[width=\linewidth]{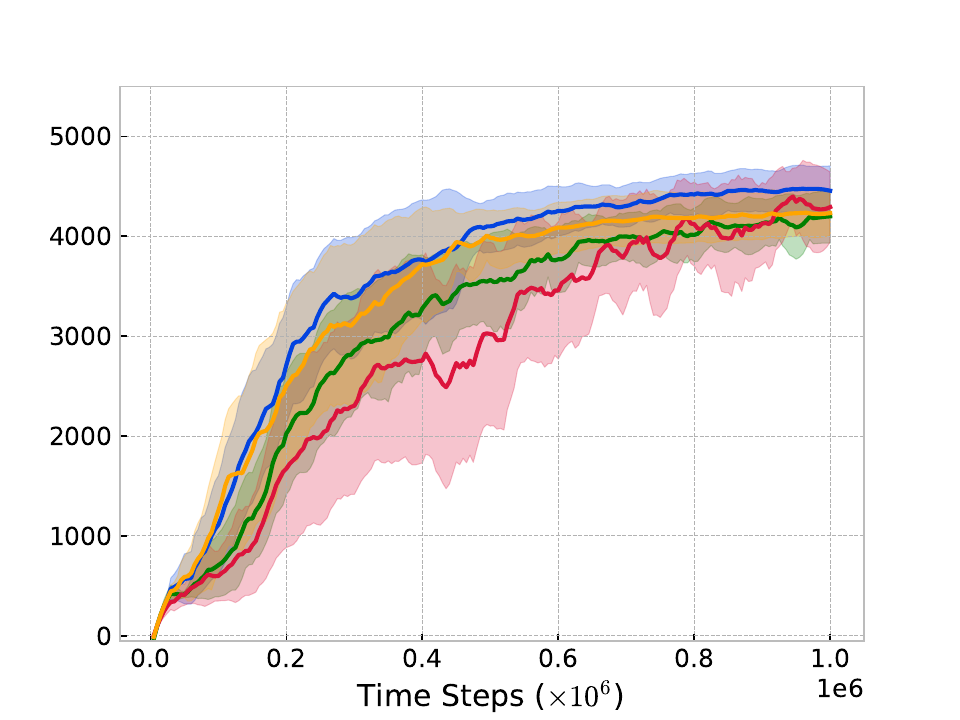}
\caption{Walker2d-v2}\label{fig:Walker2d-v2}
\end{subfigure}
\caption{Learning curves for continuous control tasks in MuJoCo over 1 million steps. Curves are averaged over 20 random seeds, where the shaded area represents the $95\%$ confidence interval of the average evaluation. All curves are smoothed with Savitzky–Golay filter for visual clarity.} \label{fig:mujoco result}
\end{center}
\end{figure}

\begin{figure}[h] 
\begin{subfigure}{0.325\textwidth}
\includegraphics[width=\linewidth]{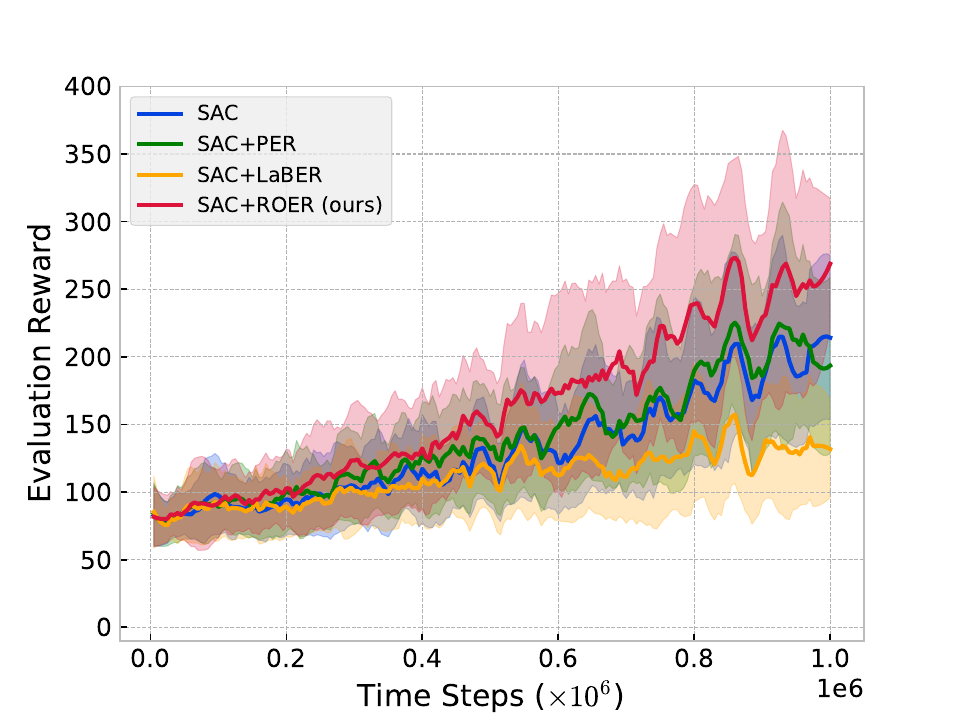}
\caption{Fish-Swim}\label{fig:fish-swim}
\end{subfigure}
\begin{subfigure}{0.325\textwidth}
\includegraphics[width=\linewidth]{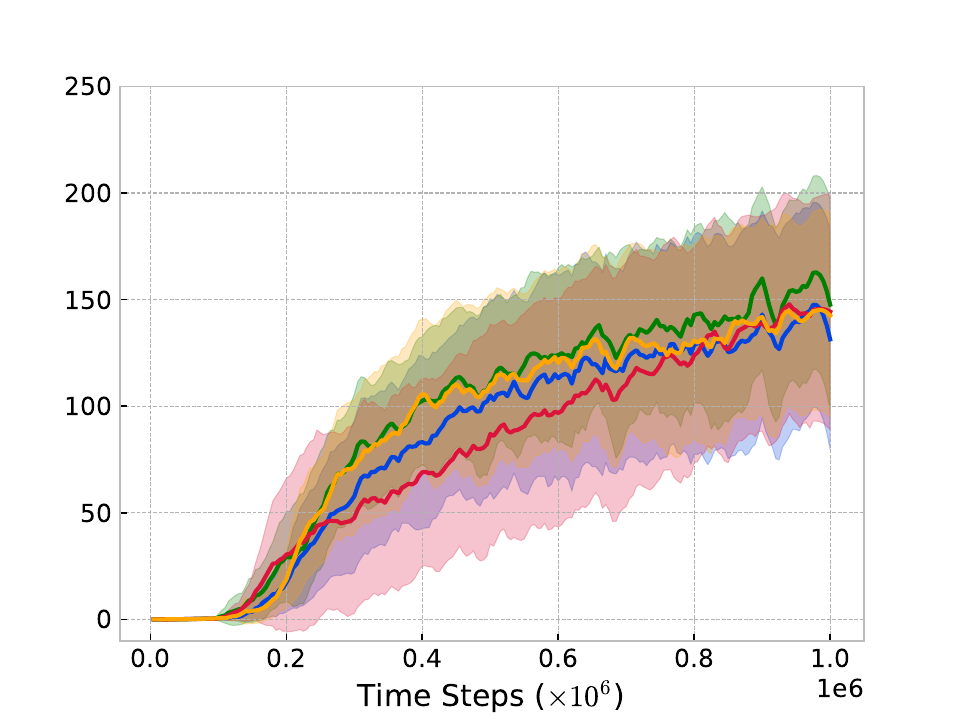}
\caption{Hopper-Hop}\label{fig:hopper-hop}
\end{subfigure}
\begin{subfigure}{0.325\textwidth}
\includegraphics[width=\linewidth]{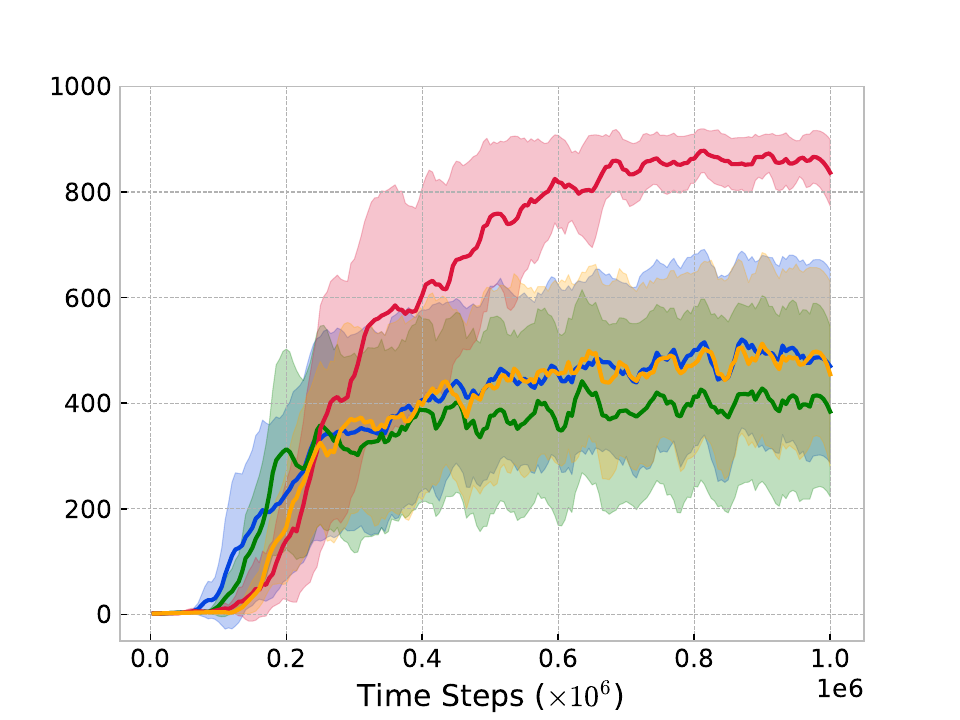}
\caption{Hopper-Stand}\label{fig:hopper-stand}
\end{subfigure}

\begin{subfigure}{0.325\textwidth}
\includegraphics[width=\linewidth]{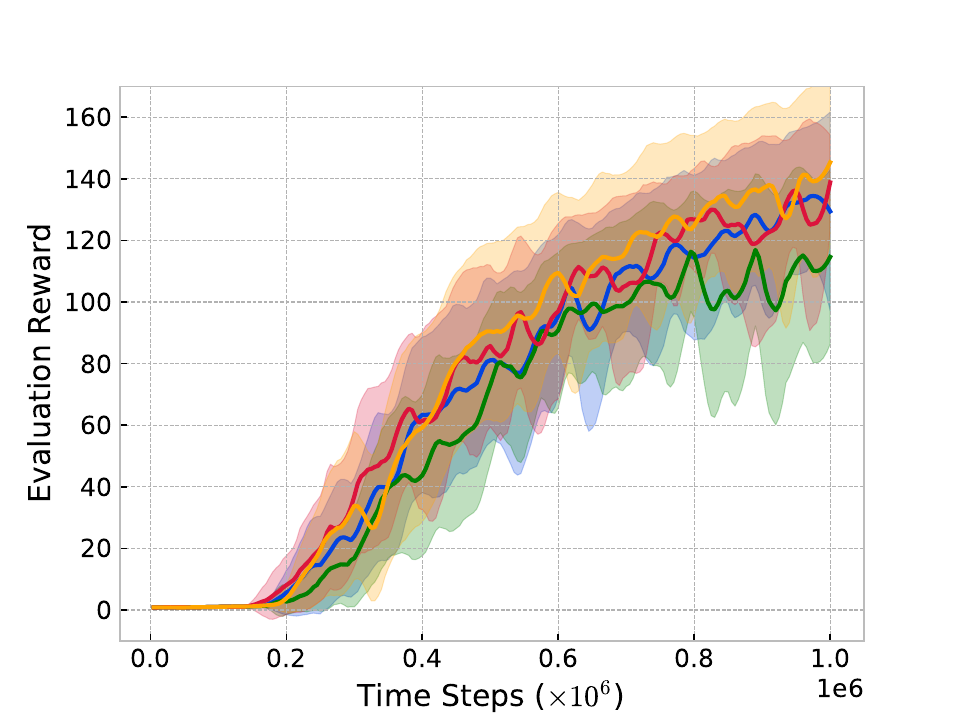}
\caption{Humanoid-Run}\label{fig:humanoid-run}
\end{subfigure}
\begin{subfigure}{0.325\textwidth}
\includegraphics[width=\linewidth]{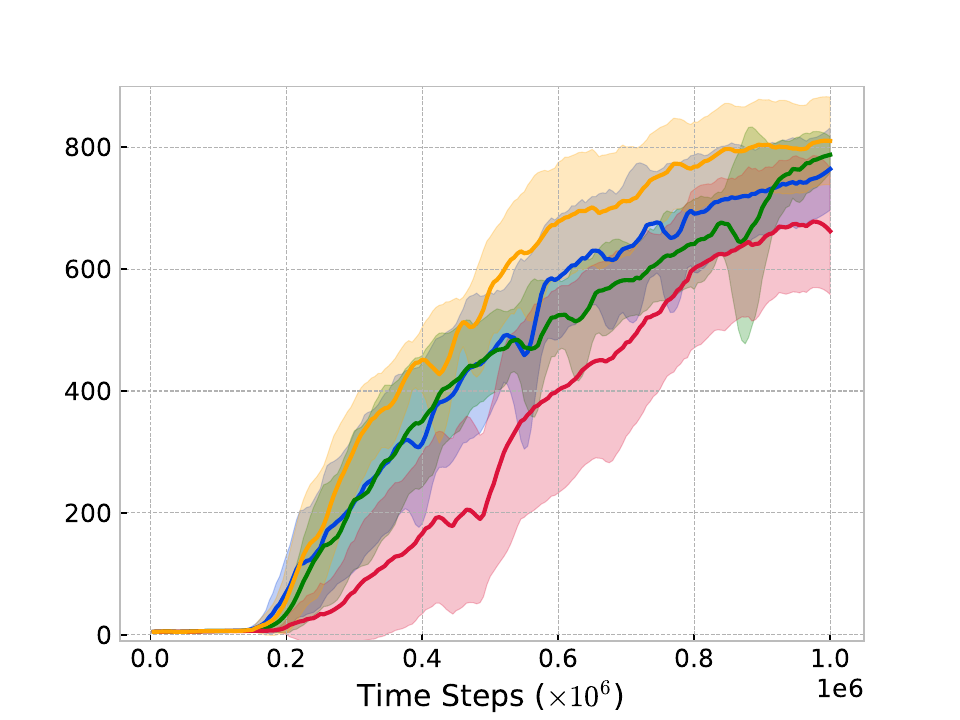}
\caption{Humanoid-Stand}\label{fig:humanoid-stand}
\end{subfigure}
\begin{subfigure}{0.325\textwidth}
\includegraphics[width=\linewidth]{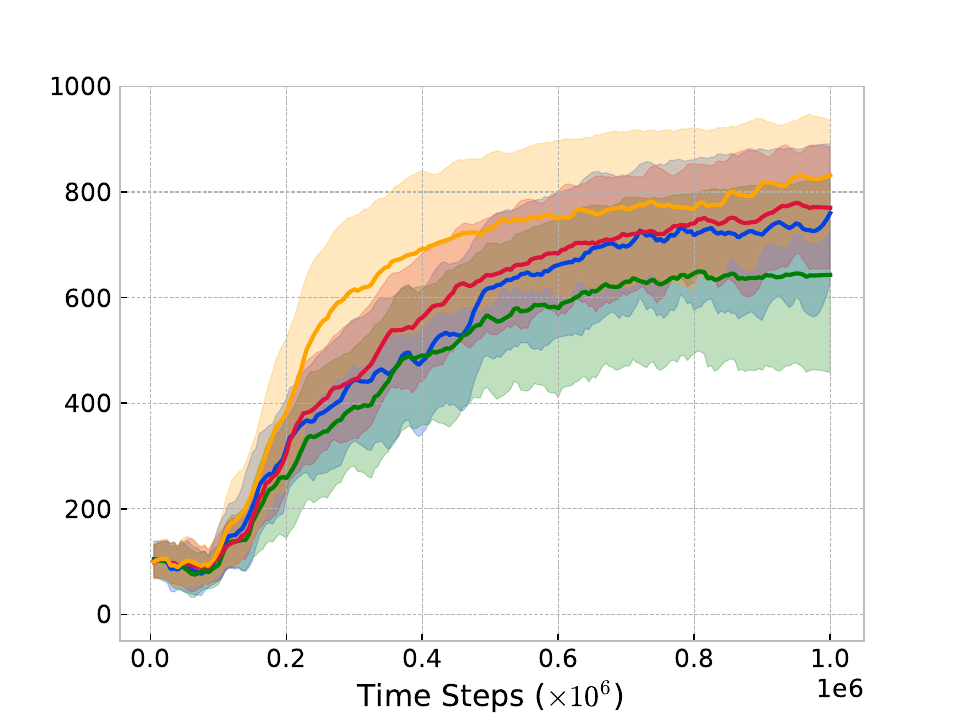}
\caption{Quadruped-Run}\label{fig:quadruped-run}
\end{subfigure}
\caption{Learning curves for continuous control tasks in DM Control over 1 million steps. Curves are averaged over 20 random seeds, where the shaded area represents the $95\%$ confidence interval of the average evaluation. All curves are smoothed with Savitzky–Golay filter for visual clarity.} \label{fig:dmc result}
\end{figure}

\begin{figure}[h] 
\begin{center}
\begin{subfigure}{0.325\textwidth}
\includegraphics[width=\linewidth]{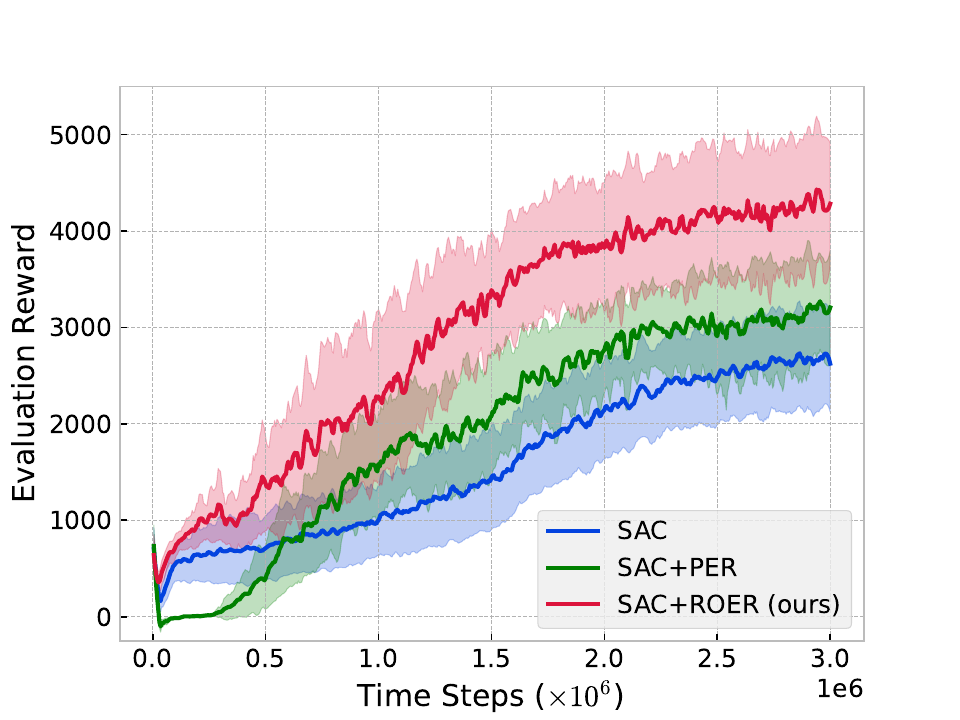}
\caption{Ant-v2}\label{fig:Ant-v2 3M}
\end{subfigure}
\begin{subfigure}{0.325\textwidth}
\includegraphics[width=\linewidth]{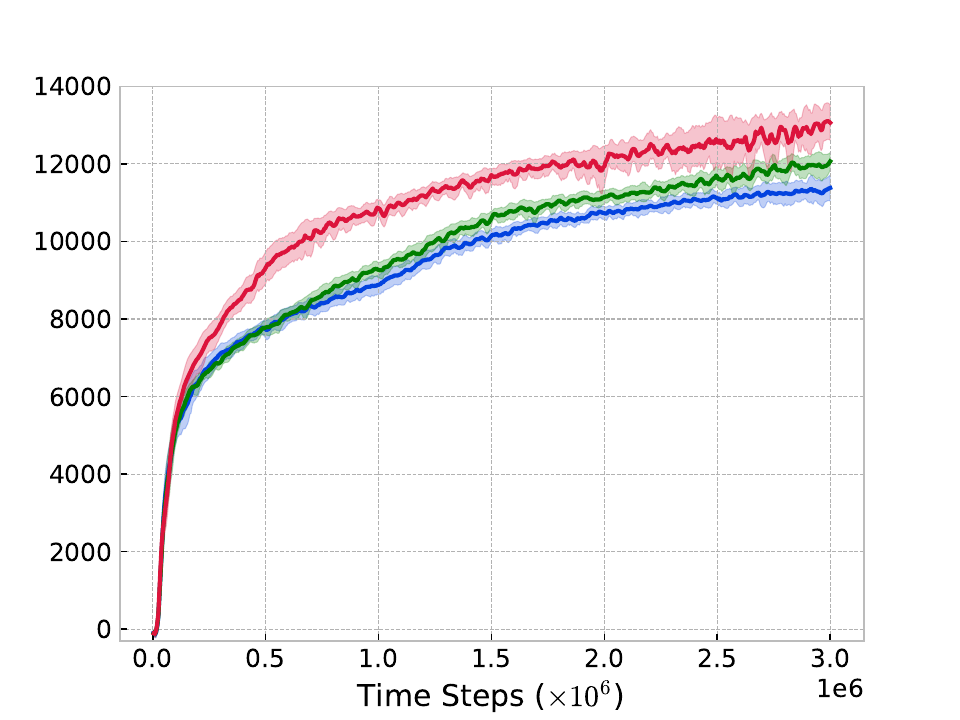}
\caption{HalfCheetah-v2}\label{fig: HalfCheetah-v2 3M}
\end{subfigure}
\begin{subfigure}{0.325\textwidth}
\includegraphics[width=\linewidth]{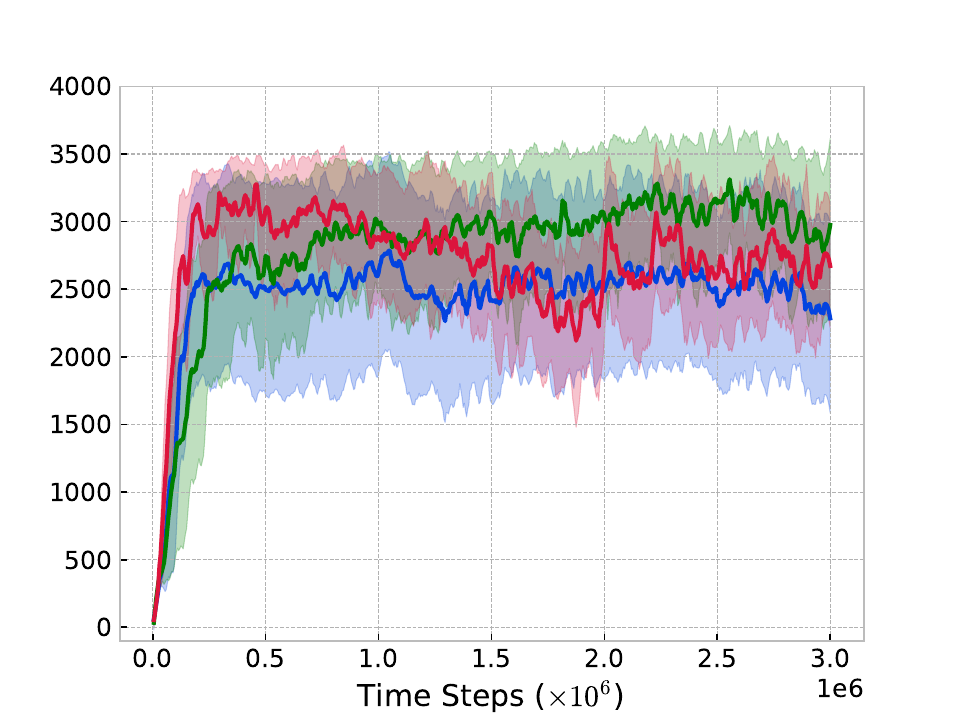}
\caption{Hopper-v2}\label{fig:Hopper-v2 3M}
\end{subfigure}

\begin{subfigure}{0.325\textwidth}
\includegraphics[width=\linewidth]{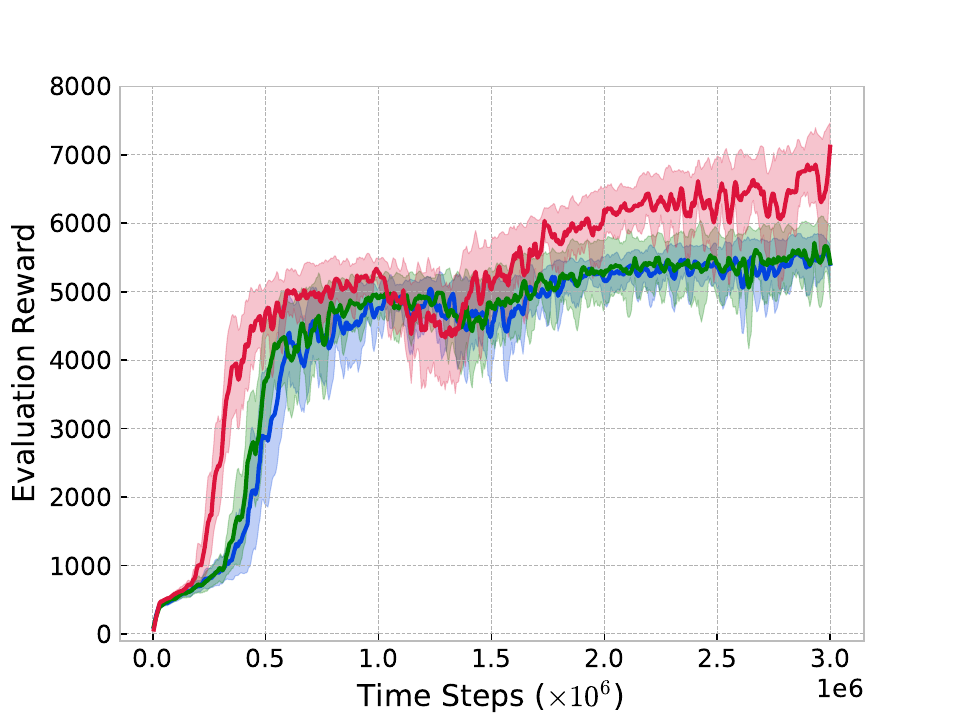}
\caption{Humanoid-v2}\label{fig:Humanoid-v2 3M}
\end{subfigure}
\begin{subfigure}{0.325\textwidth}
\includegraphics[width=\linewidth]{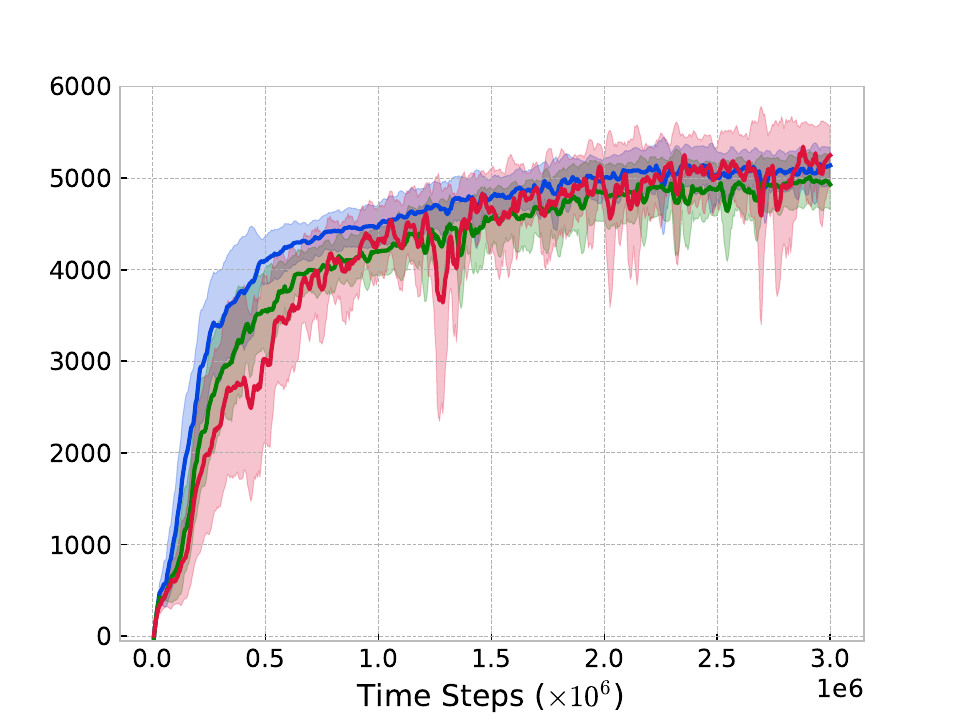}
\caption{Walker2d-v2}\label{fig:Walker2d-v2 3M}
\end{subfigure}
\caption{Learning curves for continuous control tasks in MuJoCo over 3 million steps. Curves are averaged over 10 random seeds, where the shaded area represents the $95\%$ confidence interval of the average evaluation. All curves are smoothed with Savitzky–Golay filter for visual clarity.} \label{fig:mujoco result 3M}
\end{center}
\end{figure}

In this section, we present the additional results and discussions of our experiments. The learning curves in the online setting are shown in Fig.~\ref{fig:mujoco result} for tasks in MuJoCo and Fig.~\ref{fig:dmc result} for tasks in DM Control. Besides the better performance, we note that our proposed ROER also shows faster improvement in Ant-v2, HalfCheetah-v2, Hopper-v2, Humanoid-v2, Fish-Swim and Hopper-Stand. This implies that ROER can obtain better data efficiency than the baselines. We additionally evaluate our proposed ROER in comparison to baselines in MuJoCo over 3 million steps to better illustrate the advantage of our proposed method as shown in Fig.~\ref{fig:mujoco result 3M}. ROER shows consistently better performance as in the evaluation over 1 million steps and outperforms baselines in Ant-v2, HalfCheetah-v2, and Humanoid-v2 with very little or without overlapping shaded region. We did not include LaBER for this comparison due to the long training time it takes. We note that the performance of ROER gets worse in Hopper-v2 for longer steps as it reaches the reward saturation very early and the extra training can be harmful for the policy update due to over-fitting and additional updates on $Q$-functions with weights that causes loss explosion.   

\end{document}